\documentclass[sigconf]{acmart}
\pdfoutput=1
\usepackage{lipsum}
\usepackage{mathtools}
\usepackage{cuted}
\usepackage{enumitem}
\usepackage{balance}
\usepackage{graphicx}
\usepackage{soul}
\usepackage{subcaption}
\usepackage{titlesec}
\usepackage{bm}
\usepackage{hyperref}
\pdfoutput=1
\renewcommand{\abstractname}{\uppercase{Abstract}}
\renewcommand{\refname}{\uppercase{References}}
\titleformat{\section}
  {\normalfont\Large\bfseries\uppercase}
  {\thesection}{1em}{}

\AtBeginDocument{%
  }
\usepackage[super]{nth}
\setcopyright{acmcopyright}
\copyrightyear{2024}
\acmYear{2024}
\setcopyright{rightsretained}
\acmConference[CIKM '24]{Proceedings of the 33rd ACM International
Conference on Information and Knowledge Management}{October 21--25,2024}{Boise, ID, USA}
\acmBooktitle{Proceedings of the 33rd ACM International Conference on
Information and Knowledge Management (CIKM '24), October 21--25, 2024,
Boise, ID, USA}
\acmDOI{10.1145/3627673.3679836}
\acmISBN{979-8-4007-0436-9/24/10}
\makeatletter
\gdef\@copyrightpermission{
  \begin{minipage}{0.3\columnwidth}
 \href{https://creativecommons.org/licenses/by/4.0/}{\includegraphics[width=0.90\textwidth]{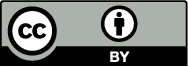}}
  \end{minipage}
  \hfill
  \begin{minipage}{0.7\columnwidth}
\href{https://creativecommons.org/licenses/by/4.0/}{This work is licensed under a Creative Commons Attribution International 4.0 License.}
  \end{minipage}
  \vspace{5pt}
}
\makeatother

\newcommand{\modelfull}{\emph{Attentive Recommendation with Contrasted Intents (ARCI)}}
\newcommand{\modeln}{ARCI}

\usepackage{titlesec}

\titlespacing*{\subsubsection}{0pt}{\parskip}{-\parskip}
\begin{document}

\title{Contrastive Learning on Medical Intents for Sequential Prescription Recommendation}
\author{Arya Hadizadeh Moghaddam}
\email{a.hadizadehm@ku.edu}
\affiliation{%
  \institution{University of Kansas}
  \city{Lawrence}
  \state{KS}
  \country{USA}
}

\author{Mohsen Nayebi Kerdabadi}
\email{mohsen.nayebi@ku.edu}
\affiliation{%
  \institution{University of Kansas}
  \city{Lawrence}
  \state{KS}
  \country{USA}
}

\author{Mei Liu}
\email{mei.liu@ufl.edu}
\affiliation{%
  \institution{University of Florida}
  \city{Gainesville}
  \state{FL}
  \country{USA}
}

\author{Zijun Yao}
\email{zyao@ku.edu}
\authornote{Corresponding author.}
\affiliation{%
  \institution{University of Kansas}
  \city{Lawrence}
  \state{KS}
  \country{USA}
}

\renewcommand{\shortauthors}{Hadizadeh Moghaddam et al.}
\setlength{\abovedisplayskip}{3pt} 
\setlength{\belowdisplayskip}{3pt} 

\setlength{\abovecaptionskip}{3pt}
\setlength{\belowcaptionskip}{-3pt}

\begin{abstract}
Recent advancements in sequential modeling applied to Electronic Health Records (EHR) have greatly influenced prescription recommender systems. 
While the recent literature on drug recommendation has shown promising performance, the study of discovering a diversity of coexisting temporal relationships at the level of medical codes over consecutive visits remains less explored. The goal of this study can be motivated from two perspectives. First, there is a need to develop a sophisticated sequential model capable of disentangling the complex relationships across sequential visits. Second, it is crucial to establish multiple and diverse health profiles for the same patient to ensure a comprehensive consideration of different medical intents in drug recommendation. To achieve this goal, we introduce Attentive Recommendation with Contrasted Intents (ARCI), a multi-level transformer-based method designed to capture the different but coexisting temporal paths across a shared sequence of visits. Specifically, we propose a novel intent-aware method with contrastive learning, that links specialized medical intents of the patients to the transformer heads for extracting distinct temporal paths associated with different health profiles. We conducted experiments on two real-world datasets for the prescription recommendation task using both ranking and classification metrics. Our results demonstrate that ARCI has outperformed the state-of-the-art prescription recommendation methods and is capable of providing interpretable insights for healthcare practitioners.
\end{abstract}



\maketitle

\vspace{-0.2cm}
\section{Introduction}
\begin{figure}[t]
        \centering
	\includegraphics[width= 8cm]{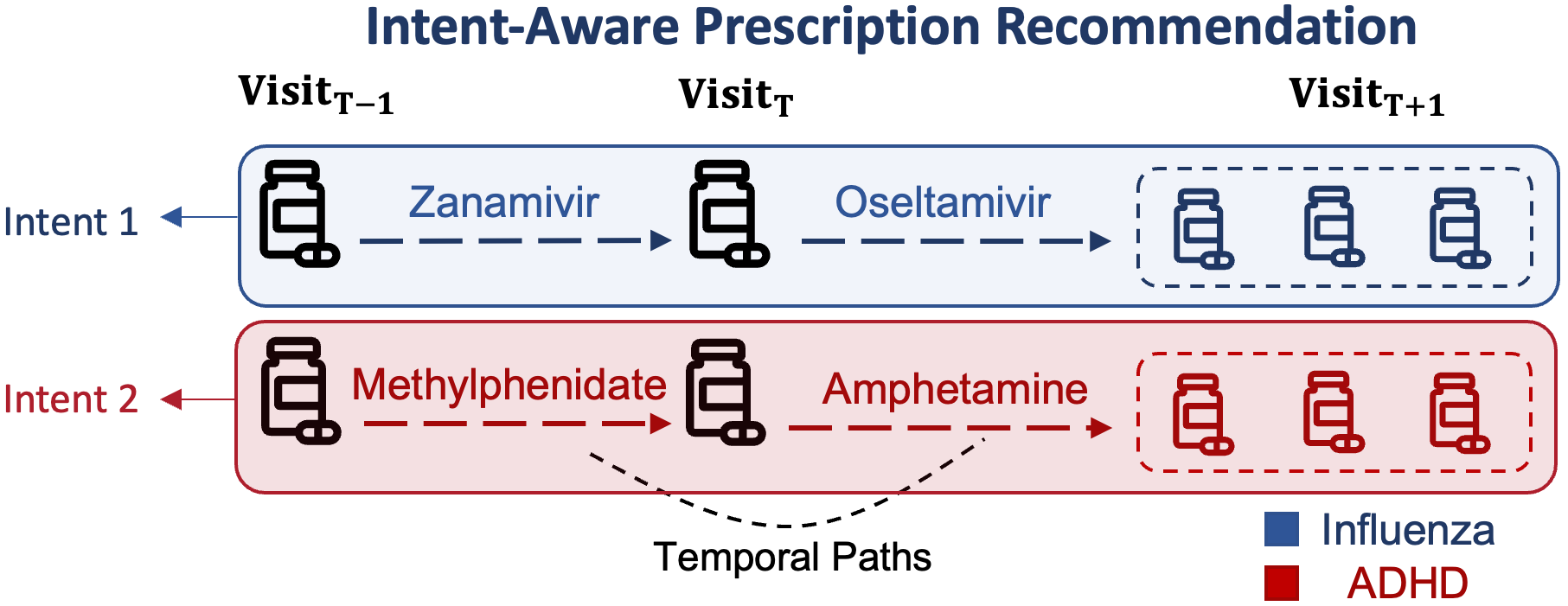}
	\caption{In EHR, a patient's sequence of visits often involves more than one thread of prescribing drugs driven by different medical intents (e.g., influenza and ADHD). We use temporal paths to distinguish the distinct dependencies among medical codes associated with different intents.}
    \label{recommenderVSmedical}
    \vspace{-0.4cm}
\end{figure}

In recent years, the abundance of Electronic Health Records (EHRs) has helped medical professionals enable a variety of data-driven applications for personalized healthcare analysis \cite{ashton2023using}. In particular, sequential prescription recommender systems have attracted a growing interest \cite{zhang2023knowledge, wu2022conditional, sun2022debiased, refine}, by supporting informed treatment decisions through the analysis of complex EHR data accumulated during the long history of individuals' medical encounters.
Aiming at predicting the medication in the next visit, a personalized algorithm is needed to examine a multitude of medical features, including diagnosis, procedures, and prescription codes, while also capturing the intricate interactive relationships among them \cite{safedrug}.

A notable challenge in building the sequential prescription recommendation arises from the multi-level structure of EHR features \cite{gct,yao2023ontology}. In EHR data, each patient consists of a sequence of healthcare visits, and within each visit, there exists a bag of medical codes for various events. Therefore, a distinctive challenge of learning temporal patterns in a sequence of medical visits is to address the heterogeneity of code interdependencies that happen simultaneously. For example, as shown in Figure \ref{recommenderVSmedical}, a patient's healthcare history can facilitate two series of drug prescriptions, addressing different treatment goals.
As a result, we believe that a single sequence of visits in EHR can be disentangled into more than one unique thread of prescribing considerations. We name each specialized prescribing thread a ``temporal path'' across consecutive visits to distinguish different types of temporal relationships. Recognizing a set of diverse temporal paths is crucial for modeling multiple concurrent medical conditions that occur in the same patient.
However, existing sequential recommender systems \cite{ranjbar2022fair, wu2019session}, such as the modeling of movie preferences, mostly study a homogeneous temporal relationship, where each ``visit'' only contains a single item at a time, making their capacity less suitable for the multi-level structure of EHR features. 

By extracting multiple dependencies over consecutive visits at the patient level, we propose to characterize temporal paths and link them to a set of distinct medical profiles (e.g., ``influenza'' and ``ADHD'' in Figure \ref{recommenderVSmedical}).
This specific design of having multiple distinct profiles in the same patient echoes a concept presented in the recent literature of recommender systems, known as ``Intent-Aware'' recommendation, which is capable of capturing diverse profiles of user preference simultaneously \cite{koniew2020classification,attmixer,chen2022intent}. For example in e-commerce recommendation, a successful intent-aware algorithm can capture a user's interest in both technology and car equipment, and by analyzing both specific interests, the system generates overall recommendations that are aligned with the user's preferences in both areas. The concept of intent in recommender systems can also be applied to EHR systems for more generalized temporal relationship learning. However, it remains a less explored research direction in prescription recommendation. A significant challenge in using intents for prescription recommendations is to ensure that each intent is able to represent a separate temporal path, and it can be clearly distinguished from other intents to represent a specialized medical profile. To achieve this goal, developing a contrastive learning \cite{chen2020simple, khosla2020supervised} framework on such medical intents constitutes a feasible approach.
Contrastive learning is generally achieved by training an objective of contrasting positive pairs (similar samples) and negative pairs (dissimilar samples) to learn a discriminative representation.
In our case, by viewing every medical intent as an anchor, the framework will encourage the recommendation models to bring the extracted temporal paths closer to the corresponding intents, meanwhile, to push each medical intent apart from the other intents in the embedding space. In this way, every intent will be maximally diverse from each other, and the specialized temporal patterns will be extracted.

To this end, in this paper, we introduce \modelfull{}, a personalized multi-level transformer-based approach designed to model simultaneous temporal paths over consecutive visits, and extract distinct medical intents for facilitating prescription recommendation with the following contributions:
\begin{itemize}[leftmargin=*,labelsep=0.5em]
    \item First, we introduce a multi-level transformer-based approach designed to extract multiple temporal paths associated with distinct medical intents, which captures both inter-visit and intra-visit medication dependencies for visit-level and cross-visit EHR representation. 
    \item Second, we propose a novel contrastive learning approach to extract diverse medical intents for regularizing the multiple temporal paths of the same patients. Each intent is linked to a dedicated transformer attention-head offering two advantages: (1) each intent and corresponding temporal path represent a specialized type of patient profile; (2) all the intents will be distinct from each other and work collectively to achieve a comprehensive patient representation.
    \item Last, we validate \modeln{} on MIMIC-III \cite{MIMIC3} and Acute Kidney Injury (AKI) \cite{liu2022development} datasets with comprehensive ranking and classification metrics on the quality of recommendations. The experimental results illustrate that the proposed method outperforms state-of-the-art recommendation approaches. Furthermore, we assess the interpretability of the proposed method to provide meaningful clinical insights for practitioners. 
\end{itemize}

\section{Methodology}

\subsection{Problem Formulation}
In the EHR data, each patient is represented as a time sequence of visits, and in each visit, various prescribed medications are recorded. Formally, for a patient $p$, there is a sequence of visit sorted by time from the earliest to the latest $\mathbf{X}^p=\langle\mathbf{x}_1, \mathbf{x}_2, \ldots, \mathbf{x}_{T}\rangle$, where $T$ represents his/her particular number of total visits\footnote{We omit the superscript $p$ for simplification when we explain for one patient.}. 
In the sequence of visits, each visit $\mathbf{x}_t$ is represented by  $\langle {m^t_1}, {m^t_2}, \dots, {m^t_{L_t}} \rangle$, a dynamic set of medication where $t$ indicates the specific visit and $L_t$ shows the total number of medication prescribed at visit $t$.\\
\textbf{Task: }Given the medication of sequential visits from $\mathbf{x}_1$ to $\mathbf{x}_{T}$ belonging to a patient, the goal of recommendation model is to predict a distinct set of medication for the next-visit at $T+1$, denotes $\langle {m^{T+1}_1}, {m^{T+1}_2}, \dots, {m^{T+1}_{L_{T+1}}} \rangle$, haven't been prescribed in earlier visits.
\begin{figure*}[ht]
        \centering
	\includegraphics[width= 0.75\textwidth]{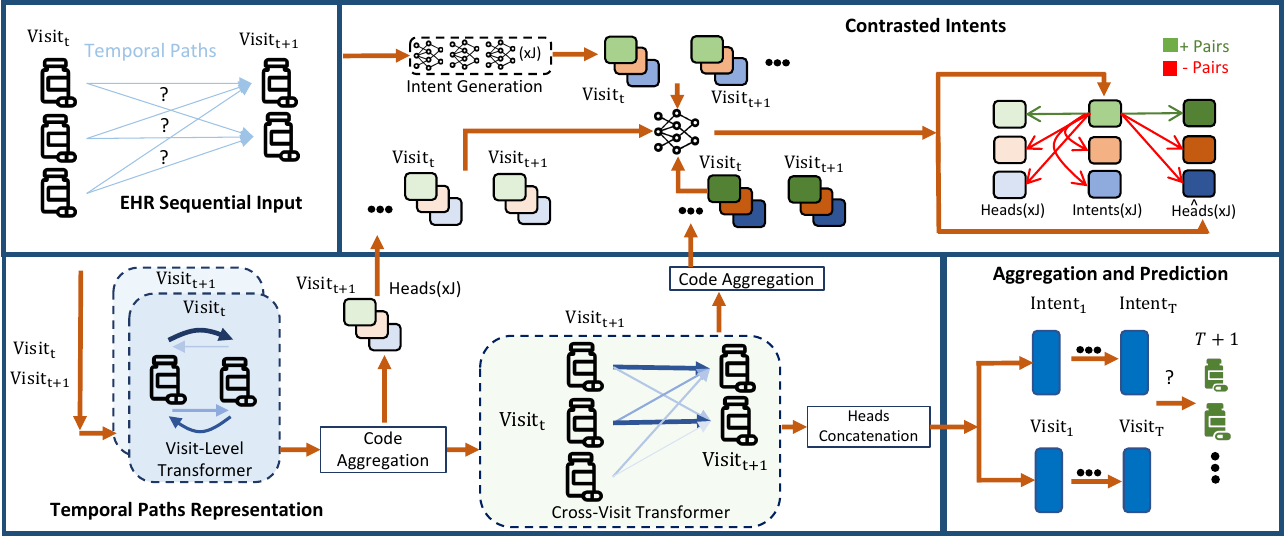}
    \caption{In the \modeln{} framework, the sequence of visits is firstly processed within the Temporal Paths Representation module including the Visit-Level Transformer to extract intra-visit dependencies, and the Cross-Visit Transformer to generate temporal paths based on inter-visit codes dependency. Meanwhile, the input embeddings of each visit are passed through multiple linear layers to generate multiple medical intents, and a contrastive learning objective is further utilized to refine the attention heads in the Contrasted Intents module. Lastly, the output of the Temporal Paths Representation module and the intent representations are fed to a visit-instance attention mechanism inside the Aggregation and Prediction module.}
    \label{method}
\end{figure*}

\subsection{Methodology Overview}
As shown in Figure \ref{method}, the proposed method consists of 
(1) \textbf{Temporal Paths Representation}, utilizing multi-level self-attention modules to generate temporal paths for enhanced embedding learning.
(2) \textbf{Contrasted Intents}, diversifying temporal paths on the same patients by linking them with different medical intents in a contrastive learning framework.
(3) \textbf{Aggregation and Prediction}, employing a visit-instance attention mechanism to prioritize visits for the final prescription prediction. 

\subsection{Temporal Paths Representation}
In this study, we adopt a multi-level transformer-based module specifically tailored for intra- and inter-visit representation.

\subsubsection{Visit-Level Transformer}\hspace{2pt}
\label{subsec:visitembed}
\\
The self-attention of Transformer plays an important role in capturing dependencies among healthcare features \cite{yang2023transformehr,gct}, and we utilize it for dependency learning between medication codes.
Firstly, the sparse input of the prescriptions is converted to a dense representation with an embedding layer in Equation \ref{eq:embd},
\begin{equation}
\label{eq:embd}
\hat{\textbf{m}}_i^t = \text{Embed}(\textbf{m}_i^t)
\end{equation} 
where $\hat{\textbf{m}_i}^t$ $\in \mathbb{R}^{d}$ is the embedding of prescription $i$ at time step $t$, $d$ is the embedding size, and $\textbf{P}_t = \langle {\hat{\textbf{m}}_1}^t, {\hat{\textbf{m}}_2}^t, ..., {\hat{\textbf{m}}_L}^t \rangle$ $\in \mathbb{R}^{B \times L_B \times d}$, where $\textbf{P}_t$ is the embeddings of prescription codes inside a visit at time $t$, $B$ is the batch size, $L_B$ is the maximum number of prescription codes within visits in the batch, and $d$ is the embedding size.
Next, the embeddings are fed into the self-attention encoder \cite{vaswani2017attention}, which focuses on extracting dependencies between prescription codes within the same visits as shown in Equation \ref{eq:visitembed}, where $\textbf{E}^j_t$ $\in \mathbb{R}^{B \times L_B \times d/J}$ represents the visit-level embedding at time step $t$ and in $j$-th head where $J$ equals to total number of heads.
\begin{equation}
\label{eq:visitembed}
\textbf{E}^j_t= \text{Self-Att}_{j}(\textbf{P}_t)
\end{equation} 

Additionally, the sequence of the embeddings for a single patient for a specific head is ${\textbf{E}^j} = \langle \textbf{E}^j_1, \textbf{E}^j_2, \dots, \textbf{E}^j_T \rangle \in \mathbb{R}^{B \times T \times L_B \times d/J}$, where $T$ is the total number of time steps.

\subsubsection{Cross-Visit Transformer}\hspace{2pt}
While the aforementioned approach enhances the prescription embedding within each single visit, it cannot fully address the cross-visit dependency learning between the codes in the consecutive time steps. Recognizing the significance of these dependencies is crucial for monitoring prescription changes over time and capturing the progression of specific conditions. Consequently, we propose a novel attention-based approach that is able to generate temporal paths over prescriptions in consecutive time steps, enabling the sharing of information between medical codes at consecutive time $t$ and $t+1$.

Self-attention generates three matrices, known as Query, Key, and Value. The first two matrices, Query and Key, are employed to construct the attention relationships between two information entities. In our study, to capture temporal paths by extracting dependencies between consecutive time steps, we adopt a new Transformer encoder where the Query is obtained from time $t$, while the Key is obtained from $t+1$ to reflect the effect of the outgoing attention and dependency from medications at $t$ to medications at $t+1$. In the formulation, the output of the visit-level code embeddings $\textbf{E}^j_t$ through each head $j$ is initially concatenated through $[\textbf{E}^1_t,...,\textbf{E}^J_t]$ $\in \mathbb{R}^{B \times L_B \times d}$, then the matrices $\hat{\textbf{Q}}_t$, $\hat{\textbf{K}}_t$, and $\hat{\textbf{V}}_t$ are derived using Equations \ref{eq:q1}, \ref{eq:k1}, and \ref{eq:v1}, where $\hat{\textbf{W}_{q}}$, $\hat{\textbf{W}_{k}}$, and $\hat{\textbf{W}_{v}}$ $\in \mathbb{R}^{d \times d}$ are learnable parameters. For the self-attention operation, the attention heads are produced as $\hat{\textbf{K}}^j_t$, $\hat{\textbf{Q}}^j_t$, and $\hat{\textbf{V}}^j_t$. 
\begin{equation}
\label{eq:q1}
\hat{\textbf{Q}}_{t}=[\textbf{E}^1_t,...,\textbf{E}^J_t] \hat{\textbf{W}}_{q}
\end{equation} 
\begin{equation}
\label{eq:k1}
\hat{\textbf{K}}_{t}=[\textbf{E}^1_t,...,\textbf{E}^J_t] \hat{\textbf{W}}_{k}
\end{equation} 
\begin{equation}
\label{eq:v1}
\hat{\textbf{V}}_{t}=[\textbf{E}^1_t,...,\textbf{E}^J_t] \hat{\textbf{W}}_{v}
\end{equation} 

We aim to find the attention between the prescriptions of time step $t$ and $t+1$, as shown in Equation \ref{eq:selfatt2}, where $\textbf{Att}^j_{t+1}$ $\in \mathbb{R}^{B \times L_{B} \times L_{B}}$ represents the cross-visit attention between consecutive in head $j$. The masking matrix $\hat{M}$ is based on the available prescription codes for the consecutive time steps as rows are associated with medications at $t+1$ and columns at $t$. The transpose function inside the $\textbf{Att}^j_{t+1}$ corresponds to the attention directed towards the prescriptions at time step $t+1$, and the sum of incoming attention for each prescription equals 1. 
\begin{equation}
\label{eq:selfatt2}
\textbf{Att}^j_{t+1}=\operatorname{softmax}((\frac{\hat{\textbf{Q}}^j_{t} ({\hat{\textbf{K}}^j}_{t+1})^T}{\sqrt{d/J}} + \hat{M})^T)
\end{equation} 

Following the generation of temporal paths, the embedding of the prescriptions at time step $t$ is shared to $t+1$ based on their cross-visit attention. Mathematically, based on Equation \ref{eq:m1} the temporal embedding is extracted, where ${\hat{\textbf{e}}^j}_{t+1}$ $\in \mathbb{R}^{B \times L_B \times d/J}$. This represents the output embedding of prescriptions at $t+1$, which is extracted from the embeddings and corresponding attentions for prescriptions at $t$.
The Cross-Visit Transformer embedding at head $j$ related to temporal paths for all time steps equals to ${\hat{\textbf{e}}^j} =  \langle\hat{\textbf{e}}^j_1, \hat{\textbf{e}}^j_2, \dots, \hat{\textbf{e}}^j_T \rangle \in \mathbb{R}^{B \times T \times L_B \times d/J}$.
\begin{equation}
\label{eq:m1}
{\hat{\textbf{e}}^j}_{t+1}=\textbf{Att}^j_{t+1} {\hat{\textbf{V}}^j}_{t}
\end{equation}
Notably, each head $j$ generates a new set of temporal embeddings between the visits. These embeddings are subsequently used for medical intents in the contrastive learning approach (Section \ref{subsec:contrastint}).

The extracted temporal path representation ${\hat{\textbf{e}}^j}_{t+1}$ is then concatenated with visit-level representation ${\hat{\textbf{V}}^j_{t+1}}$, to include both inter and intra-visit code dependencies. Following this concatenation, a linear layer is applied to derive the cross-visit embedding at $t+1$, as indicated in Equation \ref{eq:m2}, where $\textbf{W}_{T}$  $\in \mathbb{R}^{(2d/J) \times (d/J)}$.
\begin{equation}
\label{eq:m2}
{\hat{\textbf{E}}^j}_{t+1}= [{\hat{\textbf{e}}^j}_{t+1}, {\hat{\textbf{V}}^j_{t+1}}] \textbf{W}_T
\end{equation}

\begin{figure}[t]
        \centering
	\includegraphics[width= 8.5cm]{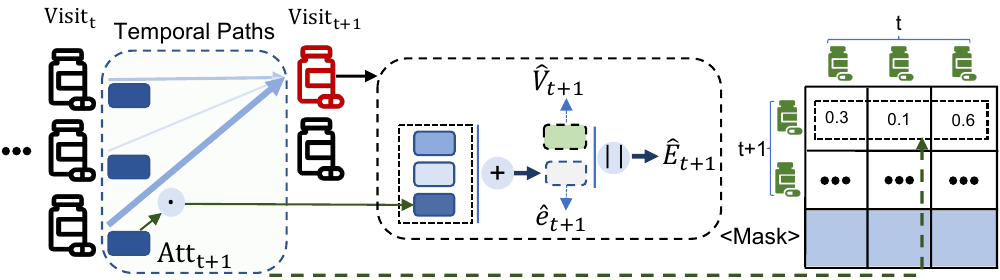}
 \vspace{0.1cm}
	\caption{The Cross-Visit Transformer generates a temporal attention matrix, where prescription embeddings at time step $t$ are multiplied by the attention matrix and shared with prescriptions at time step $t+1$. The figure illustrates the temporal attention generation for a single prescription.}\label{patientlavel}
 \vspace{-0.3cm}
\end{figure}

For further clarification, as shown in Figure \ref{patientlavel}, the attention matrix depicted in the figure is based on Equation \ref{eq:selfatt2}, where the Query is from time step $t$ and the Key from time step $t+1$, originating from the Cross-Visit Transformer architecture (Equations \ref{eq:q1}, \ref{eq:k1}, and \ref{eq:v1}). Subsequently, as illustrated in the central part of the figure, the temporal embedding for each prescription at time step $t+1$ is extracted (Equation \ref{eq:m1}), and these embeddings are then concatenated with the Value matrix from the Visit-Level Transformer output (denoted by the green box) to obtain the final embedding for prescriptions at $t+1$. (Equation \ref{eq:m2}). The attention matrix and output of the Cross-Visit Transformer represent the temporal paths between inter-visit prescriptions based on consecutive visits.

The embedding from all time steps for a batch of patients can be represented as ${\hat{\textbf{E}}^j} =  \langle \hat{\textbf{E}}^j_1, \hat{\textbf{E}}^j_2, \dots, \hat{\textbf{E}}^j_T \rangle \in \mathbb{R}^{B \times T \times L_B \times d/J}$. For $t = 1$, where incoming attention is absent, the Visit-Level Transformer output is utilized, and $\hat{\textbf{E}}^j_1 = \textbf{E}^j_1$. Notably, with this approach, the Cross-Visit Transformer at a given time $t$ takes into account the impact of prescription codes from $t=1$ to $t$.

\subsection{Contrasted Intents}
\label{subsec:contrastint}
We extracted multiple representations from attention heads in both Transformer encoders, capturing visit-level and cross-visit relationship embeddings shared among medications. Although this multi-head approach is capable of generating different temporal paths, a challenge still remains that the temporal paths associated with different attention heads are not necessarily distinct therefore may fail to represent a comprehensive range of medical intents. As shown in Figure \ref{recommenderVSmedical}, the purposes of prescribing a series of drugs are different, and each of them is associated with a specific intent. To incorporate this concept, we link each Transformer head to a specific medical intent, ensuring that multiple heads represent different considerations of a patient's health condition. Consequently, temporal paths become more medically meaningful. To achieve this, these intents and Transformer heads need to be distinct from each other to cover all the possible health profiles. Furthermore, non-distinct heads can also lead to overlapping embedding spaces, resulting in suboptimal model performance \cite{michel2019sixteen}.

Based on this motivation, a form of regularization is needed to address two key challenges: (1) ensuring the heads are distinct so the model covers the comprehensive medical aspects (such as diseases related to the heart and kidney as two different organ systems) of a complex health history, and (2) maintaining the semantics in representation, where each head is associated with a specialized medical aspect, utilizing multiple medical intents to identify medications for the recommendation. 

Inspired by the application of contrastive learning in intent-aware recommender systems \cite{chen2022intent,zou2022multi} and the generalization advantage of having distinct heads \cite{zhang2023adaptive}, we propose a novel approach for intent-aware prescription recommendation systems. 
First, 
we need to formulate the intents that can be extracted from a patient’s entire visits.
Our approach proposes to learn the intents in a computational way. Since each head is set to represent a unique intent, the number of intents equals to the number of heads. We will define $J$ linear layer blocks, each outputs a general embedding based on the patient profile as an intent. The input of these linear layers is $\hat{\textbf{P}}_t \in \mathbb{R}^{B \times d}$ which is the overall embedding obtained by summing up all the code embeddings in $\textbf{P}_{t}$. The output, as shown in Equation \ref{eq:intent}, are intent representations where $j \in \{1, 2, ..., J$\}, and $\textbf{W}^j_I \in \mathbb{R} ^{d \times d/J}$. Every intent from all time steps for a single patient can be represented as ${\textbf{Intent}^j} =  \langle \textbf{Intent}^j_1, \textbf{Intent}^j_2, \dots, \textbf{Intent}^j_T \rangle \in \mathbb{R}^{B \times T \times d/J}$.
\begin{equation}
\label{eq:intent}
{\textbf{Intent}}^j_{t}= \hat{\textbf{P}}_{t} \textbf{W}^j_I
\end{equation} 

Visit-Level and Cross-Visit Transformers share the same head index to represent the same intents that will be distinct from others. We employ a contrastive learning \cite{chen2020simple} framework. First, the prescription codes embedding for ${\textbf{E}^j}$ and ${\hat{\textbf{e}}^j}$ are summed up over individual codes respectively, so that ${\textbf{O}^j}$ and ${\hat{\textbf{O}}^j}$ $\in \mathbb{R}^{B \times T \times d/J}$ are obtained. Second, the embeddings are transformed using a linear layer as indicated in Equations \ref{eq:intent_anchor}, \ref{eq:visit_anchor}, and \ref{eq:patient_anchor}, where ${\textbf{W}}_{CL} \in \mathbb{R} ^{dT/J \times d/J}$, and $M_{P}$ is the mask to address variable number of visits.
\begin{align}
\label{eq:intent_anchor}
{\textbf{Z}}^j_{I}&= ({\textbf{Intent}^j} + M_{P}) \textbf{W}_{CL}\\
\label{eq:visit_anchor}
{\textbf{Z}}^j_{O}&= ({\textbf{O}^j} + M_{P}) \textbf{W}_{CL}\\
\label{eq:patient_anchor}
\hat{\textbf{Z}}^j_{O}&= ({\hat{\textbf{O}}^j} + M_{P}) \textbf{W}_{CL}
\end{align} 

For contrastive loss, each intent should be similar to a distinct transformer's head while being contrasted with other heads. Consequently, positive pairs are the intents and heads with the same head index $j$, while the heads and intents with different $j$ are considered negative pairs. Hence, the contrastive loss is computed according to Equations \ref{eq:visit_cl}, \ref{eq:patient_cl}, and \ref{eq:cl} for both Transformers.
\begin{equation}
\label{eq:visit_cl}
\mathcal{L}^i_{CL}=-\log \frac{\overbrace{e^{\mathrm{s}\left({\textbf{Z}}^i_{I}, {\textbf{Z}}^i_{O}\right) / \tau}}^{\text{Positive Pair}}}{\underbrace{e^{\mathrm{s}\left({\textbf{Z}}^i_{I}, {\textbf{Z}}^i_{O}\right) / \tau}}_{\text {Positive Pair }}+\underbrace{\sum_{k \neq i} e^{\mathrm{s}\left({\textbf{Z}}^i_{I}, {\textbf{Z}}^k_{O}\right) / \tau}}_{\text {Intent and Visit-Level}} + I^i}
\end{equation} 
\begin{equation}
\label{eq:patient_cl}
\hat{\mathcal{L}}^i_{CL}=-\log \frac{\overbrace{e^{\mathrm{s}\left({\textbf{Z}}^i_{I}, \hat{\textbf{Z}}^i_{O}\right) / \tau}}^{\text{Positive Pair}}}{\underbrace{e^{\mathrm{s}\left({\textbf{Z}}^i_{I}, \hat{\textbf{Z}}^i_{O}\right) / \tau}}_{\text {Positive Pair}}+\underbrace{\sum_{k \neq i} e^{\mathrm{s}\left({\textbf{Z}}^i_{I}, \hat{\textbf{Z}}^k_{O}\right) / \tau}}_{\text {Intent and Cross-Visit}}+ I^i}
\end{equation} 
\begin{equation}
\label{eq:cl}
\mathbf{\mathcal{L}}_{I}= \sum^{J}_{i = 1} {\mathcal{L}^i_{CL}} + {\hat{\mathcal{L}}^i_{CL}}
\end{equation} 
In these equations, $I^i = \sum_{k \neq i} e^{\mathrm{s}\left(\textbf{Z}^i_{I}, \textbf{Z}^K_{I}\right) / \tau}$ considers the different intents as negative pairs,  $\mathbf{\mathrm{s}}$ denotes the cosine similarity function, $\tau$ shows the temperature value.
$\mathcal{L}^i_{CL}$ indicates the contrastive learning loss for intent $i$ as the anchor and Visit-Level Transformer heads, while ${\hat{\mathcal{L}}^i_{CL}}$ represents the contrastive loss for the Cross-Visit Transformer heads. Finally, $\mathcal{L}_{I}$ is the overall contrastive loss by having all intents as anchors.

\subsection{Aggregation and Prediction}
The multi-level Transformers and intents enhance embeddings by integrating temporal and visit-level information, considering various medical aspects. Although these embeddings extract inter and intra-visit dependency among prescriptions, they do not explicitly identify which visit is more important compared to others. To identify the most crucial visits for both intents and patient embeddings, we leverage an interpretable visit-instance attention mechanism \cite{retain}. Firstly, the matrices ${\hat{\textbf{E}}^j}_t$, are concatenated based on their heads, and the representation of the codes are summed up to obtain $\hat{\textbf{E}}_t$ $\in \mathbb{R}^{B \times d}$. Subsequently, $\hat{\textbf{E}}_t$ is fed into a GRU layer, indicated in Equation \ref{eq:visit_gru}. Following the GRU layer, attention is calculated, and the final embedding is obtained using Equations \ref{eq:gru_l1}, \ref{eq:gru_softmax}, and \ref{eq:gru_l2}.
\begin{equation}
\label{eq:visit_gru}
\bm{a}_1, \bm{a}_{2}, \ldots, \bm{a}_T=\operatorname{GRU}_I\left(\mathbf{\hat{E}}_1, \mathbf{\hat{E}}_{2}, \ldots, \mathbf{\hat{E}}_T\right)
\end{equation} 
\begin{equation}
\label{eq:gru_l1}
{\bm{e}_i}= \bm{a}_{i} \textbf{W}_{a}
\end{equation} 
\begin{equation}
\label{eq:gru_softmax}
\mathbf{\alpha}_1, \mathbf{\alpha}_2, \ldots, \mathbf{\alpha}_i=\operatorname{Softmax}\left(\bm{e}_1, \bm{e}_2, \ldots, \bm{e}_i\right)
\end{equation} 
\begin{equation}
\label{eq:gru_l2}
\bm{Y}_P=\sum_{i=1}^T {\alpha_i} \odot \mathbf{\hat{E}}_{i}
\end{equation} 

A similar process is applied for intents shown in Equations \ref{eq:intent_gru}, \ref{eq:gru_l1_intent}, \ref{eq:gru_softmax_intent}, and \ref{eq:gru_l2_intent}.
\vspace{-3pt}
\begin{equation}
\label{eq:intent_gru}
\bm{b}_1, \bm{b}_{2}, \ldots, \bm{b}_T=\operatorname{GRU}_{\hat{I}}\left(\mathbf{\hat{Intent}}_1, \mathbf{\hat{Intent}}_{2}, \ldots, \mathbf{\hat{Intent}}_T\right)
\end{equation} 
\begin{equation}
\label{eq:gru_l1_intent}
{\bm{q}_i}= \bm{b}_{i} \textbf{W}_{b}
\end{equation} 
\begin{equation}
\label{eq:gru_softmax_intent}
\mathbf{\beta_1}, \mathbf{\beta_2}, \ldots, \mathbf{\beta_i}=\operatorname{Softmax}\left(\bm{q}_1, \bm{q}_2, \ldots, \bm{q}_i\right)
\end{equation} 
\begin{equation}
\label{eq:gru_l2_intent}
\bm{Y}_I=\sum_{i=1}^T {\beta_i} \odot \mathbf{Intent}_{i}
\end{equation} 
Finally, the method predicts the output using Equation \ref{eq:output}, where $\textbf{W}_{IP} \in \mathbb{R}^{2d \times d}$ and $\textbf{W}_{output} \in \mathbb{R}^{d \times H}$. $H$ is the total number of drugs that are available.
\begin{equation}
\label{eq:output}
\hat{\boldsymbol{y}}=([\bm{Y}_P, \bm{Y}_I] \textbf{W}_{IP}) \textbf{W}_{output}
\end{equation} 
\subsection{Loss Function}
The proposed method aims to predict the last visit's prescriptions based on the previous sequence, making it a multi-label classification task. In the training process, two types of loss functions are employed. Firstly, as indicated in Equation \ref{eq:bce}, Binary Cross Entropy (BCE) is used to predict the probability of each drug, 
\begin{equation}
\label{eq:bce}
\mathcal{L}_{BCE}=-\sum_{i=1}^H \boldsymbol{y}_{(i)} \log \sigma\left(\hat{\boldsymbol{y}}_{(i)} \right)+\left(1-\boldsymbol{y}_{(i)} \right) \log \left(1-\sigma\left(\hat{\boldsymbol{y}}_{(i)} \right)\right)\end{equation}
where $\boldsymbol{y}$ corresponds to the true label, $\sigma$ refers to the Sigmoid function, and $\hat{\boldsymbol{y}}_{(i)}$ refers to the prediction score for $i^{th}$ drug.

Furthermore, for enhanced result robustness and to maintain a one-margin superiority of truth labels over others, we employ the Multi-Label Hinge Loss, as identified in Equation \ref{eq:mlml}.

\begin{equation}
\label{eq:mlml}
\mathcal{L}_{m u l t i}=\sum_{i, j:{\boldsymbol{y}}_{(i)}=1, {\boldsymbol{y}}_{(j)}=0}\frac{\max \left(0,1-\left(\sigma(\hat{{\boldsymbol{y}}}_{(i)})-{\sigma({\hat{\boldsymbol{y}}}_{(j)})}\right)\right)}{|H|}
\end{equation}

Moreover, the contrastive loss ${\mathcal{L}}_{I}$ calculated in Section \ref{subsec:contrastint} is added to $\mathcal{L}_{BCE}$ and $\mathcal{L}_{m u l t i}$. Therefore, the final loss is presented in Equation \ref{eq:lloss}, where the constants $\gamma$ adapt the contrastive loss's influence and $\lambda$ is for the multi-label hinge loss. In optimization, the Adam optimizer \cite{kingma2014adam} is employed.
\begin{equation}
\label{eq:lloss}
\mathcal{L} = \mathcal{L}_{BCE} + \gamma \mathcal{L}_{I} + \lambda \mathcal{L}_{m u l t i}
\end{equation}

\section{Evaluation}
\subsection{Datasets}
In this research, two real-world datasets are utilized for the prescription recommendation task:

\noindent $\bullet$ \textbf{MIMIC-III:} MIMIC-III \cite{MIMIC3} is an open-access database that includes health-related data linked to more than 40,000 patients who were admitted to critical care unit from 2001 to 2012. In this study, we concentrate on patients with more than one visit to suggest prescriptions for their latest visit. Our training approach involves using data from their maximum four most recent visits as input for the model, determined through experimentation.\\
\noindent $\bullet$ \textbf{AKI:} The dataset on Acute Kidney Injury (AKI) \cite{liu2022development} contains healthcare data at the University of Kansas Medical Center (KUMC) from 2009 to 2021, and consists of over 135,000 hospitalized patients with the risk of having AKI. We include the three most recent visits for each patient where the two earlier visits are the input and the last one is the target visit for prescription recommendation.

\subsection{Experimental Setup}
In this study, we mainly focus on recommending the prescription of the next visit (information up to visit at time $T$ to predict target visit at $T+1$) which holds great utility for medical professionals. Therefore, we filtered out patients with only 1 visit. Generally, the AKI dataset encompasses a broader patient population by including non-ICU hospital admissions, therefore containing a significantly larger cohort.

To reduce noise in our dataset, we employ Anatomical Therapeutic Chemical (ATC) level 3 categorization for prescriptions, and our approach involves considering only those prescriptions that are non-repetitive across all visits. Repetitive prescriptions are often necessary for chronic conditions like heart disease or diabetes, where patients may require medication for years or even decades. While these medications are proven to be effective and safe, continually recommending the same treatment at every visit doesn't pose a significant challenge. Additionally, focusing solely on repetitive medication training may prioritize commonly used daily medications like painkillers and vitamins over less frequently prescribed but equally important drugs. This could limit the model's ability to understand sophisticated prescription patterns. Hence, we remove repetitive prescriptions in our main experiments.  Although our experiments primarily focused on non-repetitive prescriptions, we also conducted experiments with repetitive drugs for clarity and consistency with previous studies reports (Section 3.4.3).

For \modeln{} and baselines, we use a greedy search approach to find the best hyperparameters for a comprehensive evaluation. We randomly split the data into 80\% for training and 20\% for testing, and we repeat this approach 7 times and report the means and confidence intervals of results. The optimal configuration for the \modeln{} on MIMIC-III entails $J$ equals to 4, $\tau$ to 0.05, batch size to 128, $\lambda$ to 0.05, $\gamma$ to 0.2, and $d$ to 256. For AKI, $J$ set to 6, batch size to 128, $\tau$ to 0.05, $\lambda$ to 0.05, $\gamma$ to 0.1, and $d$ to 256.

The statistics of the datasets are shown in Table \ref{Statistical}. In this project, the programming language is Python, and we utilize PyTorch \cite{pytorch} for the method architecture. Furthermore, \modeln{} is compatible with \textit{PyHealth} \cite{pyhealth} framework, and we employ its baseline implementations. We release the project source code\footnote{\url{https://github.com/aryahm1375/ARCI}.} on GitHub.
\begin{table}[t]
\centering
\addtolength{\tabcolsep}{-0.2em}
\caption{Statistics of the datasets.}
\label{Statistical}
\small
\makebox[\linewidth]{
    \begin{tabular}{@{}ccc@{}}
    \hline
    & \multicolumn{1}{c}{\bf{MIMIC-III}} & \multicolumn{1}{c}{\bf{AKI}}\\
        \hline
    Avg. \# of drugs per visit  & 10.86 & 8.28  \\
    \# of target drugs  & 191 & 222  \\
    \# of target visits        & 10104 & 63940  \\
    \# of patients & 6442 & 63940  \\
    Avg. \# of input visits & 2.52 & 2 \\
    \hline
    \end{tabular}
}
\vspace{-0.1cm}
\end{table}
\begin{table*}[ht]
\renewcommand*{\arraystretch}{1.2}
\footnotesize
\centering
\addtolength{\leftskip}{-1cm}

\caption{Performance comparison on MIMIC-III based on ranking metrics (Hit@K and NDCG@K). The reported values include means and 95\% confidence interval calculated from folds.}
\label{rankingmimicIII}
\begin{tabular}{cccccc}
\hline
\textbf{Method}       & \textbf{Hit \& NDCG@1}$\uparrow$             & \textbf{Hit@3}$\uparrow$              & \textbf{NDCG@3}$\uparrow$             & \textbf{Hit@5}$\uparrow$              & \textbf{NDCG@5}$\uparrow$             \\
\hline
\textbf{Dr. Agent}       & 0.416 $\pm$ 0.008 & 0.171 $\pm$ 0.005 & 0.443 $\pm$ 0.005 & 0.097 $\pm$ 0.004 & 0.488 $\pm$ 0.003 \\
\textbf{Retain}      & \underline{0.459 $\pm$ 0.009} & \underline{0.194 $\pm$ 0.006} & \underline{0.493 $\pm$ 0.006} & \underline{0.112 $\pm$ 0.005} & \underline{0.534 $\pm$ 0.005}  \\
\textbf{Micron}      & 0.333 $\pm$ 0.005 & 0.123 $\pm$ 0.007   & 0.365 $\pm$ 0.005  & 0.062 $\pm$ 0.005 & 0.410 $\pm$ 0.004 \\
\textbf{SafeDrug}    & 0.350 $\pm$ 0.014 & 0.130 $\pm$ 0.004  & 0.377 $\pm$ 0.012 & 0.059 $\pm$ 0.004 & 0.411 $\pm$ 0.012 \\
\textbf{MoleRec}     & 0.336 $\pm$ 0.017 & 0.122 $\pm$ 0.012 & 0.369 $\pm$ 0.015 & 0.067 $\pm$ 0.006 & 0.414 $\pm$ 0.014 \\
\textbf{GAMENet}     & 0.409 $\pm$ 0.009 & 0.164 $\pm$ 0.005 & 0.441 $\pm$ 0.007 & 0.089 $\pm$ 0.005 & 0.484 $\pm$ 0.005 \\

\hline
\textbf{Transformer} & 0.416 $\pm$ 0.010 & 0.172 $\pm$ 0.009 & 0.448 $\pm$ 0.008 & 0.099 $\pm$ 0.006 & 0.493 $\pm$ 0.007  \\
\textbf{\modeln{} w/o intent}& 0.462 $\pm$ 0.011 & 0.196 $\pm$ 0.005 & 0.489 $\pm$ 0.007 & 0.117 $\pm$ 0.007 & 0.532 $\pm$ 0.004  \\
\textbf{\modeln{} w intent w/o}  $\mathbf{\mathcal{L}}_{I}$ & 0.475 $\pm$ 0.010  & 0.192 $\pm$ 0.007 & 0.495 $\pm$ 0.007 & 0.113 $\pm$ 0.008 & 0.538 $\pm$ 0.006 \\
\textbf{\modeln{}}  & \textbf{0.491 $\pm$ 0.011} & \textbf{0.209 $\pm$ 0.008}  & \textbf{0.516 $\pm$ 0.008} & \textbf{0.121 $\pm$ 0.010} & \textbf{0.556 $\pm$ 0.006}\\
\hline

\end{tabular}
\end{table*}

\begin{table*}[ht]
\renewcommand*{\arraystretch}{1.2}
\footnotesize
\centering
\addtolength{\leftskip}{-1cm}
\caption{Performance comparison on AKI based on ranking metrics (Hit@K and NDCG@K). The reported values include means and 95\% confidence interval calculated from folds.}
\label{table:rankingaki}
\begin{tabular}{cccccc}
\hline
\textbf{Method}            & \textbf{Hit\& NDCG@1}$\uparrow$          & \textbf{Hit@3}$\uparrow$            & \textbf{NDCG@3}$\uparrow$            & \textbf{Hit@5}$\uparrow$             & \textbf{NDCG@5}$\uparrow$           \\
\hline
\textbf{Dr. Agent}        & \underline{0.407 $\pm$ 0.003} & \underline{0.299 $\pm$ 0.002} & \underline{0.551 $\pm$ 0.003} & \underline{0.218 $\pm$ 0.003} & \underline{0.608 $\pm$ 0.002}  \\
\textbf{Retain}      & 0.398 $\pm$ 0.002 & 0.294 $\pm$ 0.004 & 0.546 $\pm$ 0.002 & 0.213 $\pm$ 0.002 & 0.606 $\pm$ 0.002 \\
\textbf{Micron}      & 0.383 $\pm$ 0.008 & 0.263 $\pm$ 0.010 & 0.526 $\pm$ 0.007 & 0.182 $\pm$ 0.008  & 0.587 $\pm$ 0.005 \\
\textbf{SafeDrug}    & 0.396 $\pm$ 0.003  & 0.277 $\pm$ 0.004 & 0.535 $\pm$ 0.004 & 0.198 $\pm$ 0.005 & 0.590 $\pm$ 0.005 \\
\textbf{MoleRec}      & 0.380 $\pm$ 0.007 & 0.261 $\pm$ 0.010 & 0.525 $\pm$ 0.007  & 0.183 $\pm$ 0.006 & 0.587 $\pm$ 0.006  \\
\textbf{GAMENet}     & 0.394 $\pm$ 0.004 & 0.274 $\pm$ 0.007 & 0.536 $\pm$ 0.006 & 0.192 $\pm$ 0.007 & 0.593 $\pm$ 0.005\\
\hline
\textbf{Transformer} & 0.383 $\pm$ 0.003 & 0.280 $\pm$ 0.005 & 0.526 $\pm$ 0.003 & 0.200 $\pm$ 0.004  & 0.585 $\pm$ 0.003 \\
\textbf{\modeln{} w/o intent}      & 0.408 $\pm$ 0.001 & 0.294 $\pm$ 0.002  & 0.553 $\pm$ 0.002 & 0.212 $\pm$ 0.003 & 0.612 $\pm$ 0.002  \\
\textbf{\modeln{} w intent w/o} $\mathbf{\mathcal{L}}_{I}$ & 0.412 $\pm$ 0.002  & 0.303 $\pm$ 0.004 & 0.560 $\pm$ 0.004   & 0.220 $\pm$ 0.002 & 0.622 $\pm$ 0.003 \\
\textbf{\modeln{}}  & \textbf{0.419 $\pm$ 0.003}  & \textbf{0.308 $\pm$ 0.002} & \textbf{0.569 $\pm$ 0.002} & \textbf{0.225 $\pm$ 0.003}  & \textbf{0.631 $\pm$ 0.002}\\
\hline
\end{tabular}
\vspace{-1mm}
\end{table*}
\begin{table*}[ht]
\renewcommand*{\arraystretch}{1.2}
\footnotesize
\centering
\caption{Performance comparison on MIMIC-III and AKI based on classification metrics (PRAUC, F1 Score, and Jaccard) and drug safety metric (DDI). The reported values include means and 95\% confidence interval calculated from folds.}
\label{table:classification}
\setlength{\tabcolsep}{2pt}

\begin{tabular}{@{}ccccc|cccc@{}}
\hline
& \multicolumn{4}{c}{\bf{MIMIC-III}} & \multicolumn{4}{|c}{\bf{AKI}} \\
\hline
\textbf{Method}      & \textbf{PRAUC}$\uparrow$               & \textbf{F1}$\uparrow$                  & \textbf{Jaccard}$\uparrow$       & \textbf{DDI}$\downarrow$        & \textbf{PRAUC}$\uparrow$               & \textbf{F1}$\uparrow$                  & \textbf{Jaccard}$\uparrow$     & \textbf{DDI}$\downarrow$        \\
\hline

\textbf{Dr. Agent}      & 0.324 $\pm$ 0.003 & 0.212 $\pm$ 0.008 & 0.137 $\pm$ 0.005 & 0.047 $\pm$ 0.002& 0.350 $\pm$ 0.001 & 0.251 $\pm$ 0.001  & \underline{0.168 $\pm$ 0.001} & 0.074 $\pm$ 0.001\\
\textbf{Retain}      & \underline{0.367 $\pm$ 0.004} & \underline{0.263 $\pm$ 0.003}  & \underline{0.170 $\pm$ 0.002} & 0.052 $\pm$ 0.001 &\underline{0.351 $\pm$ 0.001}  & \underline{0.253 $\pm$ 0.001}  & 0.167 $\pm$ 0.001 & 0.072 $\pm$ 0.001  \\
\textbf{Micron}       & 0.273 $\pm$ 0.003 & 0.228 $\pm$ 0.005  & 0.143 $\pm$ 0.003 & \underline{0.047 $\pm$ 0.002}  & 0.329 $\pm$ 0.003 & 0.229 $\pm$ 0.003 & 0.146 $\pm$ 0.002 & \textbf{0.055 $\pm$ 0.002}  \\
\textbf{SafeDrug}    & 0.270 $\pm$ 0.007 & 0.190 $\pm$ 0.012 & 0.118 $\pm$ 0.008& 0.057 $\pm$ 0.004 & 0.336 $\pm$ 0.001  & 0.237 $\pm$ 0.002  & 0.156 $\pm$ 0.001  & 0.071 $\pm$ 0.001 \\
\textbf{MoleRec}      & 0.275 $\pm$ 0.007  & 0.228 $\pm$ 0.005 & 0.143 $\pm$ 0.003 & \textbf{0.046 $\pm$ 0.002}  & 0.328 $\pm$ 0.002 & 0.230 $\pm$ 0.001 & 0.146 $\pm$ 0.001  & \underline{0.057 $\pm$ 0.001} \\
\textbf{GAMENet}     & 0.319 $\pm$ 0.004 & 0.239 $\pm$ 0.008  & 0.153 $\pm$ 0.005 & 0.054 $\pm$ 0.003  & 0.345 $\pm$ 0.002 & 0.251 $\pm$ 0.003  & 0.164 $\pm$ 0.003 & 0.070 $\pm$ 0.002\\
\hline
\textbf{Transformer} & 0.340 $\pm$ 0.006 & 0.232 $\pm$ 0.007 & 0.150 $\pm$ 0.005 & 0.048 $\pm$ 0.001 & 0.333 $\pm$ 0.001 & 0.237 $\pm$ 0.001 & 0.156 $\pm$ 0.001 & 0.072 $\pm$ 0.001   \\
\textbf{\modeln{} w/o intent}                    & 0.352 $\pm$ 0.005 & 0.273 $\pm$ 0.005 & 0.175 $\pm$ 0.004 & 0.058 $\pm$ 0.002& 0.354 $\pm$ 0.001 & 0.268 $\pm$ 0.001 & 0.180 $\pm$ 0.001 & 0.074 $\pm$ 0.003\\
\textbf{\modeln{} w intent w/o}  $\mathbf{\mathcal{L}}_{I}$ & 0.366 $\pm$ 0.005 & 0.282 $\pm$ 0.005 & 0.183 $\pm$ 0.003 & 0.052 $\pm$ 0.001&0.360 $\pm$ 0.001 & 0.272 $\pm$ 0.001  & 0.182 $\pm$ 0.001 & 0.062 $\pm$ 0.002\\
\textbf{\modeln{}} & \textbf{0.376 $\pm$ 0.004} & \textbf{0.299 $\pm$ 0.004} & \textbf{0.194 $\pm$ 0.003} & 0.051 $\pm$ 0.001&\textbf{0.366 $\pm$ 0.001} & \textbf{0.283 $\pm$ 0.002} & \textbf{0.190 $\pm$ 0.001} & 0.063 $\pm$ 0.002\\
\hline

\end{tabular}
\end{table*}
\subsection{Evaluation Metrics}
In this research, we utilize both well-known classification and ranking metrics to evaluate \modeln{} comprehensively. For classification purposes, we employ PRAUC, F1, and Jaccard, metrics. To evaluate drug safety, we employ metrics based on drug-drug interaction (DDI) rates, and for ranking, we use Hit@K and NDCG@K with K being the rank. In the following, the formulations are presented:\\
\noindent \textbf{Jaccard} measures the similarity between two sets by calculating the ratio of the intersection to the union of ground truth, indicated by equation \ref{eq:jaccard}.
\begin{equation}
\label{eq:jaccard}
\textbf { Jaccard }=\frac{1}{\sum_p^N \sum_t^{\hat{T}_p} 1} \sum_p^N \sum_t^{\hat{T}_p} \frac{\left|y_t^{(p)} \cap \hat{Y}_t^{(p)}\right|}{\left|y_t^{(p)} \cup \hat{Y}_t^{(p)}\right|}
\end{equation} 
where, $N$ equals to the total number of patients, $\hat{T}_p$ denotes the number of visits to predict for patient $p$, ${y}_t^{(p)}$ to ground truth medications for patient $p$ at time step $t$, and $\hat{Y}_t^{(p)} = \{ \hat{y}_t^{(p)} > \theta \}$ where $\theta$ denotes a threshold, chosen through a greedy search approach from values from 0 to 1 with intervals of 0.1, relying on the performance metrics evaluated on the test dataset. \\
\noindent \textbf{F1} score is derived from Precision and Recall, where $Precision = |y_t^{(p)} \cap \hat{Y}_t^{(p)}| / |y_t^{(p)}|$ and $Recall_t^{(p)} = |y_t^{(p)} \cap \hat{Y}_t^{(p)}| / |\hat{Y}_t^{(p)}|$. The final score is obtained as indicated in equation \ref{eq:f1}.
\begin{equation}
\label{eq:f1}
\textbf{F1}=\frac{1}{\sum_p^N \sum_t^{\hat{T}_p} 1} \sum_p^N \sum_t^{\hat{T}_p} \frac{2 \times Precision_t^{(p)} \times Recall_t^{(p)}}{Precision_t^{(p)}+Recall_t^{(p)}}
\end{equation} 
\noindent  \textbf{PRAUC} quantifies the performance of a classification model by measuring the area under the precision-recall curve. The formulation of PRAUC is indicated in equation \ref{eq:prauc_final}, wherein $\operatorname{PRAUC}_t^{(p)}=\sum_{l=1}^{|\mathcal{H}|} \operatorname{Precision}(l)_t^{(p)} \Delta \operatorname{Recall}(l)_t^{(p)}$ and $\Delta \operatorname{Recall}(l)_t^{(p)}=\operatorname{Recall}(l)_t^{(p)}\\ - \hspace{2pt} \operatorname{ Recall}(l-1)_t^{(p)}$, where $l$ is the rank of the sequence of the retrieved prescription. 
\begin{equation}
\label{eq:prauc_final}
\operatorname{\textbf{PRAUC}}= \frac{1}{\Sigma_p^N \Sigma_t^{\hat{T}_p} 1} \sum_k^N \sum_t^{\hat{T}_p} \operatorname{PRAUC}_t^{(p)}
\end{equation}
\noindent \textbf{DDI }rate measures medication safety by calculating the rate at which predicted prescriptions interact with each other \cite{gamenet1}, indicated in equation \ref{DDI}, where $u_i$ equals to the $i^{th}$ prescriptions prediction output, and $\mathcal{G}_d$ is the DDI graph.\\
\begin{equation}
\label{DDI}
\textbf {DDI }=\frac{\sum_p^N \sum_t^{\hat{T}_p} \sum_{i, j}\left|\left\{\left(u_i, u_j\right) \in \hat{Y}_t^{(p)} \mid\left(u_i, u_j\right) \in \mathcal{G}_d\right\}\right|}{\sum_p^N \sum_t^{\hat{T}_p} \sum_{i, j} 1}
\end{equation}

We also utilize the ranking evaluation to discuss the method's performance in different ranks, and how the model is reliable. Notably, for all of the metrics if the specific rank is $k$ we only consider inputs with target visits that have more than $k-1$ prescriptions. \\
\noindent \textbf{Hit@k} equals the hit rate of top $k$ ranked prescriptions which all of them appear in the ground truth drugs, as indicated in equation \ref{eq:hitrate} where $Ranked(\hat{y}_t^{(p)}, k)$ equals to top $k$ ranked prescription based on their prediction score.
\begin{equation}
\label{eq:hitrate}
\small
\textbf{Hit@k}=\frac{1}{\Sigma_p^N \Sigma_t^{\hat{T}_p} 1} \sum_p^N \sum_t^{\hat{T}_p} \omega_t^{(p)};\hspace{-8pt}\quad \omega_t^{(P)}=\left\{\begin{array}{l}
\hspace{-3pt}1:Ranked(\hat{y}_t^{(p)}, k) \subset y_t^{(p)} \\
\hspace{-3pt}0:\text {otherwise }
\end{array}\right.
\end{equation}
\noindent \textbf{NDCG@k} is a commonly used metric for ranking the performance of recommendation systems. The formulation is shown in equation \ref{eq:ndcg_total} , where in $DCG^{(p)}_t@ k=\sum_{i=1}^k Rank(\sigma(\hat{y}_t^{(p)}), k)_i/\log _2(i+1)$,  $IDCG^{(p)}_t@ k=\sum_{i=1}^k Rank(y_t^{(p)}, k)_i/\log _2(i+1)$ and $Rank(y_t^{(p)}, k)$ equals to score of the prescription at rank $k$.
\begin{equation}
\label{eq:ndcg_total}
\textbf{NDCG}@ K=\frac{1}{\Sigma_p^N \Sigma_t^{\hat{T}_p} 1} \frac{DCG^{(p)}_t@ K}{IDCG^{(p)}_t@ K}
\end{equation} 
\subsection{Baselines}
We utilize the state-of-the-art healthcare predictive methods and prescription recommendation systems as baselines: (1) \noindent \textbf{Dr. Agent} \cite{gao2020dr} utilizes RNN with dual policy gradient agents and a dynamic skip connection for adaptive focus on pertinent information. (2) \noindent \textbf{Retain} \cite{retain} incorporates a dual-RNN network to capture the interpretable influence of the visits and medical features for the prediction tasks. (3) \noindent \textbf{Transformer's Encoder} \cite{vaswani2017attention}, our research utilizes the transformer's self-attention for both Visit-Level and Cross-Visit Transformers representations and it is one of the most important baselines to compare our method with. (4) \noindent \textbf{Micron} \cite{yang2021change} a drug recommendation model using residual recurrent neural networks to update changes in patient health (5) \noindent \textbf{SafeDrug} \cite{safedrug} utilizes a global message passing neural network to encode the functionality of prescriptions molecules with considering the DDI to recommend safe prescriptions (6) \noindent \textbf{MoleRec} \cite{molerec} a structure-aware encoding method that contains hierarchical architecture aimed at modeling interactions with considering DDI. (7) \noindent \textbf{GAMENet} \cite{gamenet1} employs a memory module and graph neural networks to incorporate the knowledge graph of drug-drug interactions, while representing longitudinal patient records as the query.
\subsection{Results and Discussion}
We conducted experiments on \modeln{} to address the following research questions:
\begin{itemize}[leftmargin=*,labelsep=0.5em]
\item \textbf{RQ1} What is the performance of \modeln{} compared to other state-of-the-art healthcare predictive methods and prescription recommendation systems?
\item \textbf{RQ2} How differently do various components of \modeln{} contribute to the output results?
\item \textbf{RQ3} How does the \modeln{} perform in the recommendation task with accounting for repetitive prescriptions?
\item \textbf{RQ4} How does the proposed method take into account interpretability?

\end{itemize}
\begin{figure}[ht]
  \setlength{\abovecaptionskip}{5pt}

  \centering
  \begin{subfigure}[t]{0.23\textwidth}
    \includegraphics[width=\textwidth]{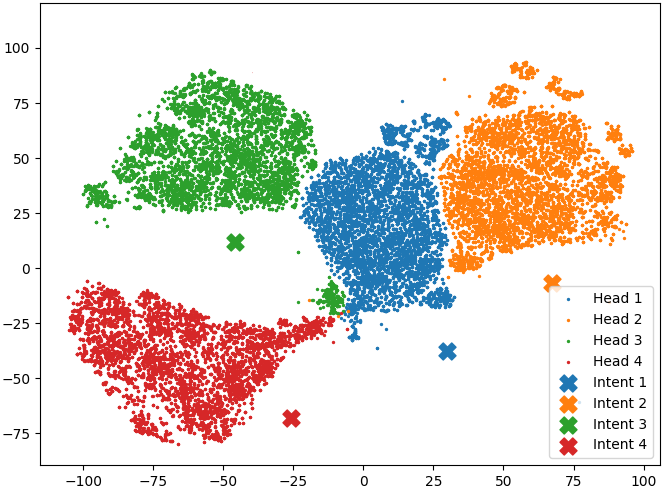}
    \caption{MIMIC-III}
  \end{subfigure}
 \begin{subfigure}[t]{0.23\textwidth}
    \includegraphics[width=\textwidth]{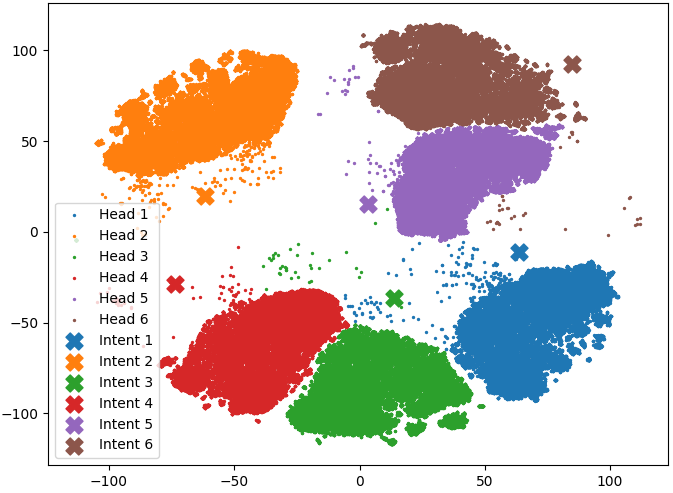}
    \caption{AKI}
  \end{subfigure}
  \vspace{0.1cm}
  \caption{t-SNE of the heads and intents. For better illustration, only the means of the intents' representations are shown as crosses. The figure depicts how different heads are successfully clustered into different groups that are aligned with their associated intents.}
  \label{TSNE}
\vspace{-10pt}
\end{figure}
\begin{figure}[t]
  \centering
    \setlength{\abovecaptionskip}{5pt}
  \begin{subfigure}[t]{0.23\textwidth}
    \includegraphics[width=\textwidth]{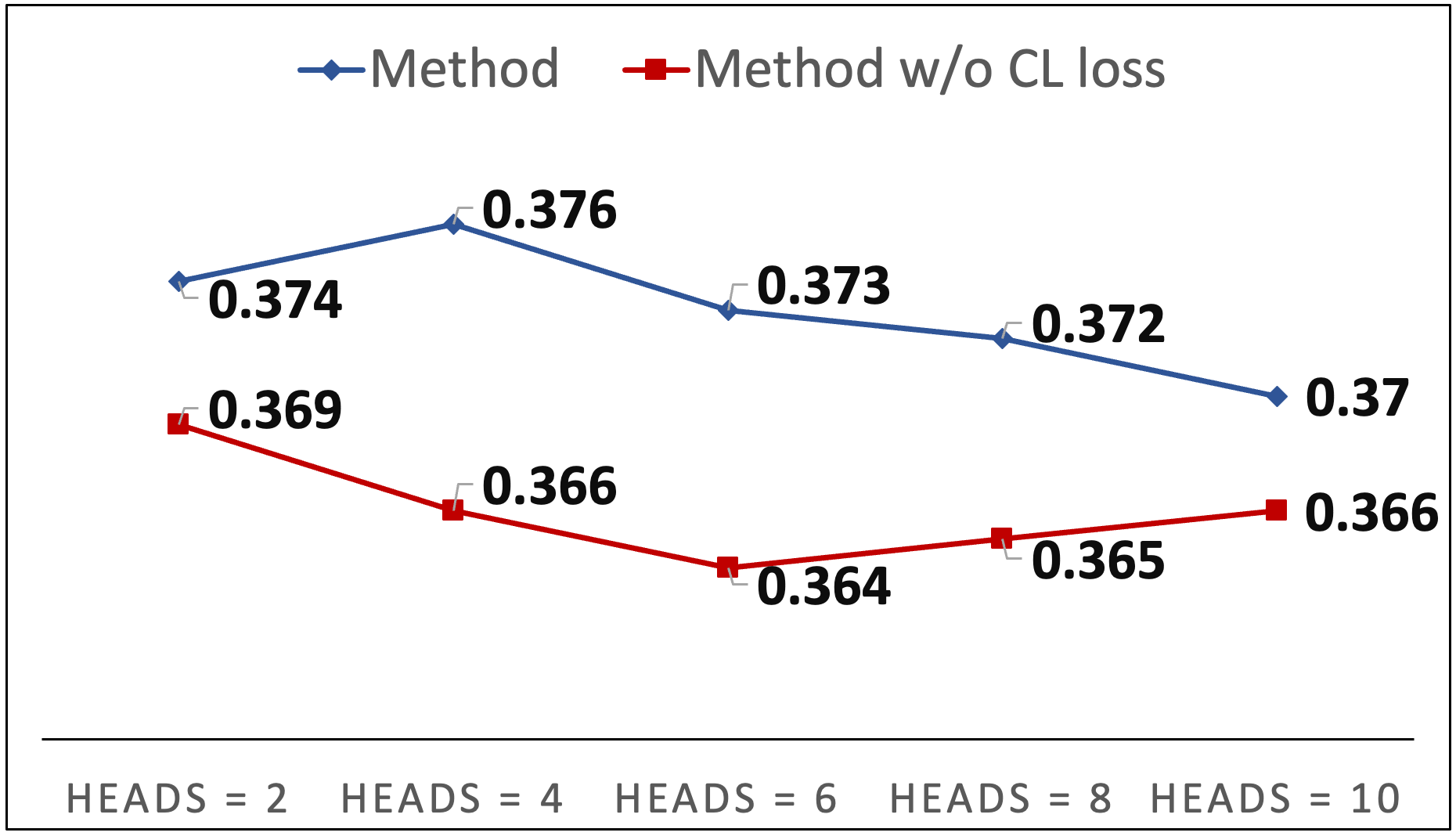}
    \caption{MIMIC-III}
  \end{subfigure}
 \begin{subfigure}[t]{0.23\textwidth}
    \includegraphics[width=\textwidth]{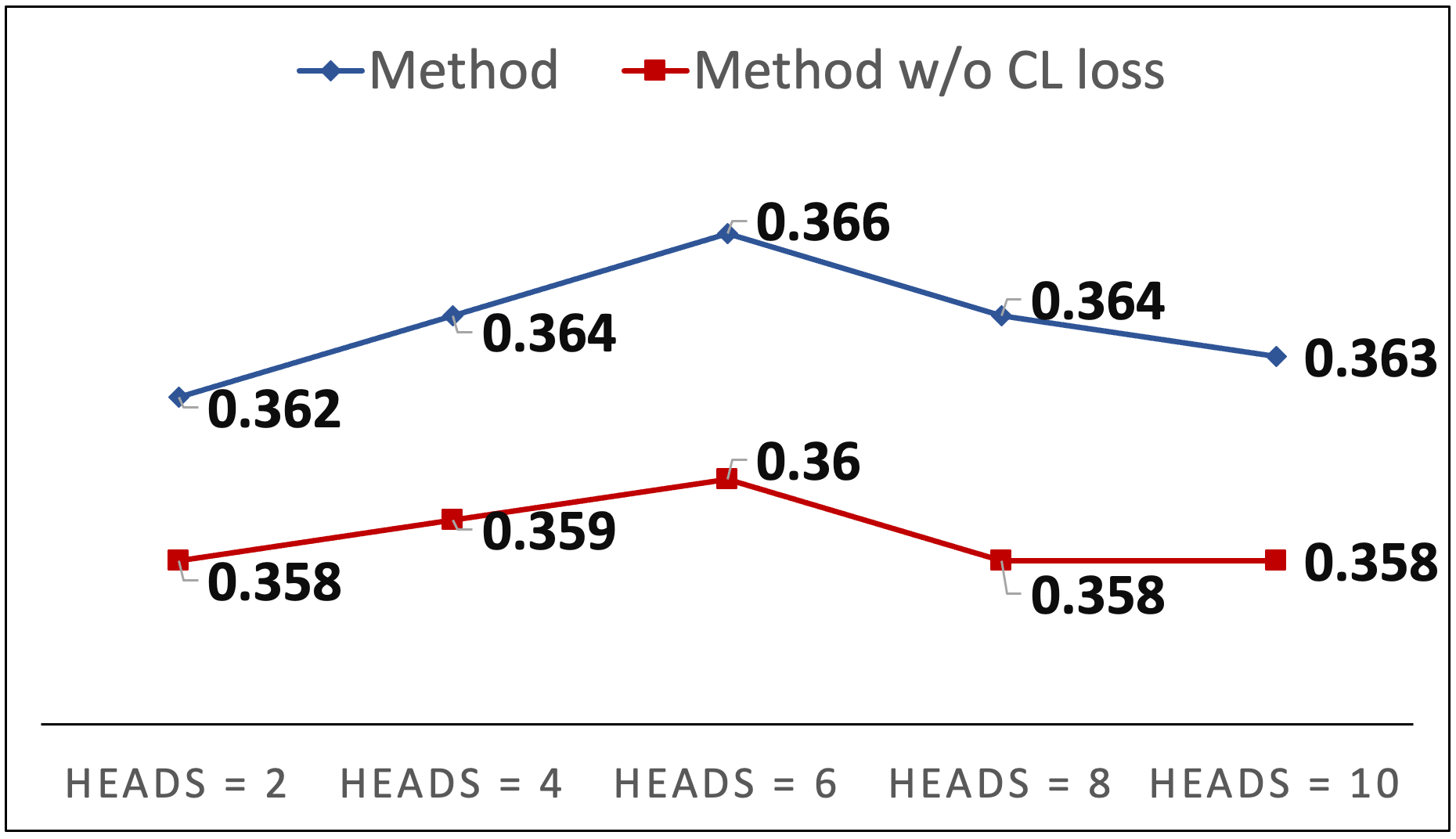}
    \caption{AKI}
  \end{subfigure}
  \vspace{0.1cm}
  \caption{The effect of varying numbers of heads and intents on the PRAUC Metric with and without $\mathcal{L}_{I}$ on the MIMIC-III and AKI datasets.}
  \label{CLfig}
\end{figure}

\begin{figure}[t]
  \setlength{\abovecaptionskip}{5pt}

  \centering
  \begin{subfigure}[t]{0.21\textwidth}
    \includegraphics[width=\textwidth]{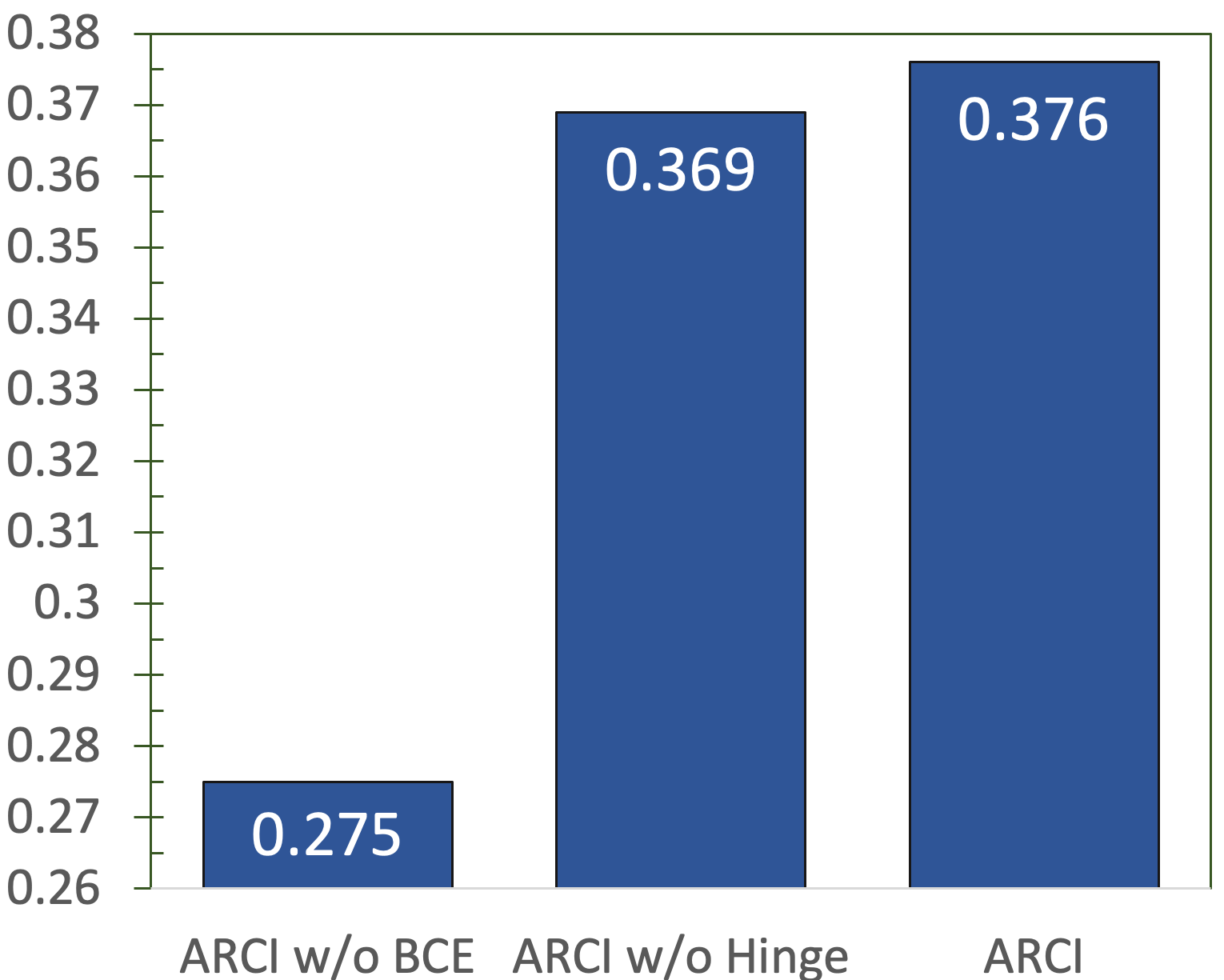}
    \caption{MIMIC-III}
  \end{subfigure}
 \begin{subfigure}[t]{0.21\textwidth}
    \includegraphics[width=\textwidth]{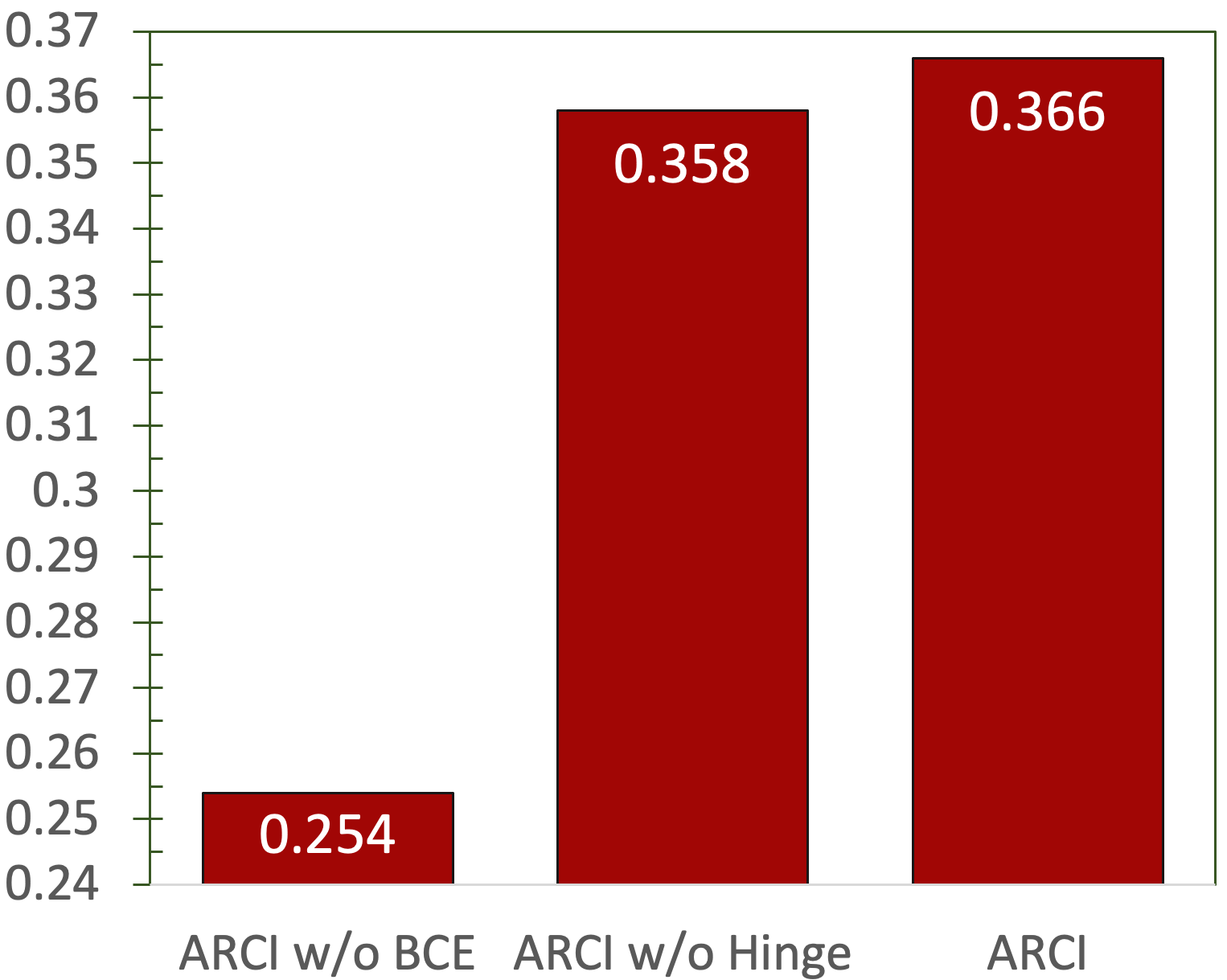}
    \caption{AKI}
  \end{subfigure}
  \vspace{0.1cm}
  \caption{The impact of removing various loss functions on the PRAUC metric for the MIMIC-III and AKI datasets.}
  \label{losses}
  \vspace{-5pt}
\end{figure}

\begin{table*}[t]
\vspace{8pt}
\caption{Performance comparison on MIMIC-III and AKI based on classification metrics and DDI including repetitive prescriptions. The reported values include means and 95\% confidence interval calculated from folds.}
\footnotesize
\begin{tabular}{@{}ccccc|cccc@{}}
\hline
\setlength{\tabcolsep}{2pt}
& \multicolumn{4}{c}{\bf{MIMIC-III}} & \multicolumn{4}{|c}{\bf{AKI}} \\
\hline
\textbf{Method}      & \textbf{PRAUC}$\uparrow$               & \textbf{F1}$\uparrow$                  & \textbf{Jaccard}$\uparrow$       & \textbf{DDI}$\downarrow$        & \textbf{PRAUC}$\uparrow$               & \textbf{F1}$\uparrow$                  & \textbf{Jaccard}$\uparrow$     & \textbf{DDI}$\downarrow$        \\
\hline
\textbf{Dr. Agent}   & 0.688 $\pm$ 0.003 & 0.565 $\pm$ 0.007 & 0.405 $\pm$ 0.007 & 0.079 $\pm$ 0.005 & 0.432 $\pm$ 0.002 & 0.315 $\pm$ 0.003 & 0.214 $\pm$ 0.002 & 0.082 $\pm$ 0.002 \\
\textbf{Retain}      & 0.686 $\pm$ 0.002 & 0.565 $\pm$ 0.004 & 0.406 $\pm$ 0.003 & 0.082 $\pm$ 0.001 & 0.449 $\pm$ 0.001 & 0.321 $\pm$ 0.002 & 0.219 $\pm$ 0.002 & 0.083 $\pm$ 0.001 \\
\textbf{Micron}      & 0.686 $\pm$ 0.002 & 0.566 $\pm$ 0.003 & 0.406 $\pm$ 0.003 & \textbf{0.069 $\pm$ 0.002} & 0.433 $\pm$ 0.001   & 0.331 $\pm$ 0.002 & 0.229 $\pm$ 0.001 & \underline{0.071 $\pm$ 0.001} \\
\textbf{SafeDrug} & 0.711 $\pm$ 0.002 & 0.590 $\pm$ 0.002 & 0.428 $\pm$ 0.002 & 0.074 $\pm$ 0.003 & 0.450 $\pm$ 0.001 & 0.326 $\pm$ 0.002 & 0.222 $\pm$ 0.002 & 0.075 $\pm$ 0.002 \\
\textbf{MoleRec}  & \underline{0.714 $\pm$ 0.002} & \underline{0.595 $\pm$ 0.002} & \underline{0.435 $\pm$ 0.002} & \underline{0.070 $\pm$ 0.001} & \underline{0.453 $\pm$ 0.001}   & \underline{0.332 $\pm$ 0.001} & \underline{0.231 $\pm$ 0.001}   & \textbf{0.068 $\pm$ 0.001} \\
\textbf{GAMENet}     & 0.706 $\pm$ 0.003 & 0.580 $\pm$ 0.005 & 0.419 $\pm$ 0.005 & 0.078 $\pm$ 0.002 & 0.451 $\pm$ 0.001 & 0.328 $\pm$ 0.001 & 0.230$\pm$ 0.002  & 0.076 $\pm$ 0.001 \\
\textbf{Transformer} & 0.713 $\pm$ 0.003 & 0.591 $\pm$ 0.003 & 0.430 $\pm$ 0.003 & 0.079 $\pm$ 0.003 & 0.439 $\pm$ 0.001 & 0.313 $\pm$ 0.002 & 0.213 $\pm$ 0.001 & 0.078 $\pm$ 0.002 \\
\hline
\textbf{ARCI}        & \textbf{0.727 $\pm$ 0.002} & \textbf{0.606 $\pm$ 0.002} & \textbf{0.446 $\pm$ 0.002} & 0.073 $\pm$ 0.001 & \textbf{0.472 $\pm$ 0.001} & \textbf{0.341 $\pm$ 0.002} & \textbf{0.241 $\pm$ 0.002} & 0.075 $\pm$ 0.002\\
\hline
\label{repetitive}
\end{tabular}
\end{table*}

\subsubsection{RQ1. Performance Comparison with Baselines:} \hspace{2pt} Tables \ref{rankingmimicIII} and \ref{table:rankingaki} provides the comparison between baselines and \modeln{} based on the ranking metrics for MIMIC-III and AKI datasets. The results highlight superior performance of \modeln{} across all metrics, particularly demonstrating an advantage over one of the best-performing baselines, Retain, in the top rank. While Retain employs an RNN-based methodology for using both code-level and visit-level feature learning, the observed marginal difference suggests that temporal paths with contrasting patterns are more effective for temporal learning compared to interpretable RNNs. The enhancement made by \modeln{} compared to prescription recommendation systems, namely SafeDrug \cite{safedrug}, MoleRec \cite{molerec}, GAMENet \cite{gamenet1}, and Micron \cite{micron}, demonstrating that the interdependency between medical codes across consecutive time steps results in more informative embeddings for future visit prediction. 

Furthermore, we conducted experiments based on classification metrics, as outlined in Table \ref{table:classification}. The results show the improved performance of our proposed method over all baselines in terms of the classification metrics. 
MoleRec, SafeDrug, GAMENet, and Micron use DDI loss \cite{gamenet1} as part of their method. In contrast, our training process excludes this loss, with our primary focus on comprehensive embedding learning.  
Although Micron and MoleRec have better DDI rates compared to \modeln{} (0.5\% for MIMIC-III and 0.8\% for AKI), the marginal differences between the models in terms of PRAUC (10.1\% for MIMIC-III and 3.7\% for AKI) suggest that \modeln{} remains an effective prescription recommender.
\subsubsection{RQ2. Ablation Studies:} \hspace{2pt}
In the ablation studies, we investigate the impact of omitting each submodule on the overall performance. We include the Transformer's encoder in the ablation analysis, as \modeln{} is built upon the Transformer architecture. \modeln{} has two main principal contributions: (1) Extracting temporal paths using the Cross-Visit Transformer, and (2) proposing the Contrasted Intent module for linking temporal paths and dependencies with distinct health profiles and capturing diverse embeddings. Results in Tables \ref{rankingmimicIII}, \ref{table:rankingaki}, and \ref{table:classification} indicate adding Cross-Visit Transformer (\modeln{} w/o intent) enhances Transformer's performance, showing the effectiveness of temporal paths compared to treating features as a bag-of-codes. Another observation is adding linear layers (\modeln{} w intent w/o $\mathbf{\mathcal{L}}_{I}$) as intents without any contrastive loss is beneficial, showing that incorporating general patient health information, even without any regularization, has a positive impact. Finally, based on the results and margins we can assume that $\mathbf{\mathcal{L}}_{I}$ is effective in achieving more general embeddings, a conclusion supported by Figure \ref{TSNE}. In this figure, the head representations from both datasets are considerably clustered into distinct groups. Additionally, the mean of the intents' representations is close to their associated groups different from each other, showing the influence of $\mathbf{\mathcal{L}}_{I}$ in generalization.
Furthermore, we explored the impact of varying the number of heads in the output results extracted from MIMIC-III. As depicted in Figure \ref{CLfig}, the maximum effect of $\mathbf{\mathcal{L}}{I}$ occurs when $J = 4$ for MIMIC and $J = 6$ for AKI because the optimal number of intents depends on the variability of medical conditions, linked to the number of patients which suggests that AKI has a higher variability in AKI with a higher J value. For \modeln{}, we combined three different loss functions for evaluation, and the impact of the contrastive loss has been examined. In Figure \ref{losses}, we evaluate the effects of the BCE and Hinge losses. As shown in the figure, BCE is the most effective for the multi-label classification task, while the Hinge loss provides a smaller performance improvement.
\subsubsection{RQ3. Including Repetitive Prescriptions:}\hspace{2pt}
To have more consistency with the previous model's results in the literature as they include repetitive prescriptions and to have a comprehensive assessment, we extended our experiments to include repetitive prescriptions. As shown in Table \ref{repetitive}, \modeln{} has outperformed the baselines for both datasets across various classification metrics. Specifically, the best-performing baseline is the MoleRec model for the MIMIC-III and AKI datasets, which differs from the results presented in Table \ref{table:classification}. These findings suggest that the Retain model, which includes non-repetitive prescriptions, can outperform state-of-the-art prescription recommenders. However, for repetitive drugs, state-of-the-art models are more effective. The comparisons between \modeln{}, Retain, MoleRec, and SafeDrug in Tables \ref{repetitive} illustrate that the intent-aware recommender with a multi-level Transformer architecture is effective for reliable prescription recommendations in the two aforementioned settings.

\subsubsection{RQ4. Interpretability:}
\begin{figure}[t]
  \centering
  \begin{subfigure}[h]{0.23\textwidth}
    \includegraphics[width=\textwidth]{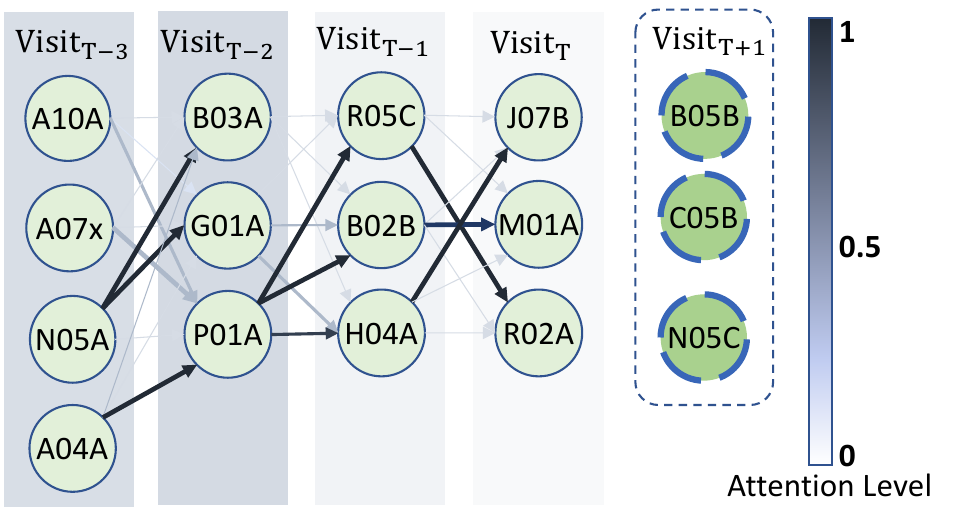}
    \caption{\nth{1} Intent Temporal Path}
  \end{subfigure}
  \vspace{3 pt}
 \begin{subfigure}[h]{0.23\textwidth}
    \includegraphics[width=\textwidth]{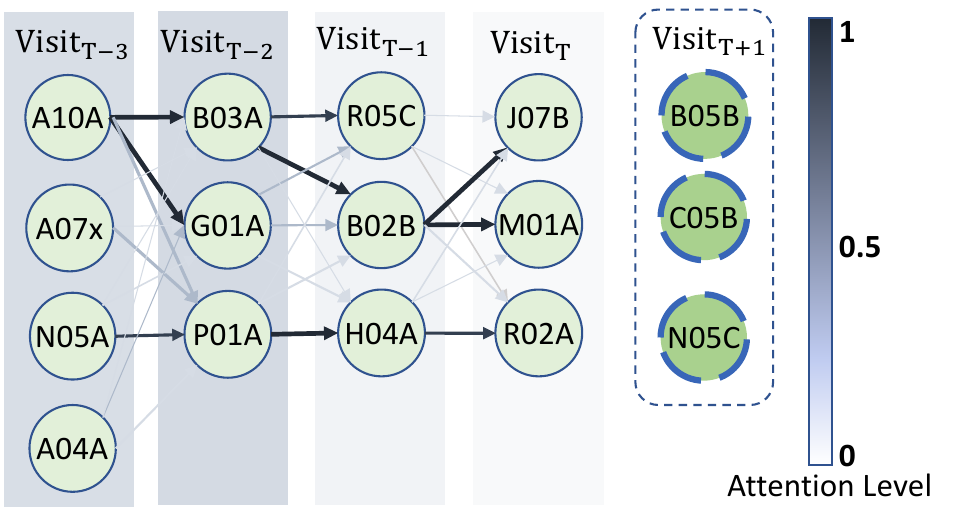}
    \caption{\nth{2} Intent Temporal Path}
  \vspace{3 pt}
  \end{subfigure}
  \begin{subfigure}[h]{0.23\textwidth}
    \includegraphics[width=\textwidth]{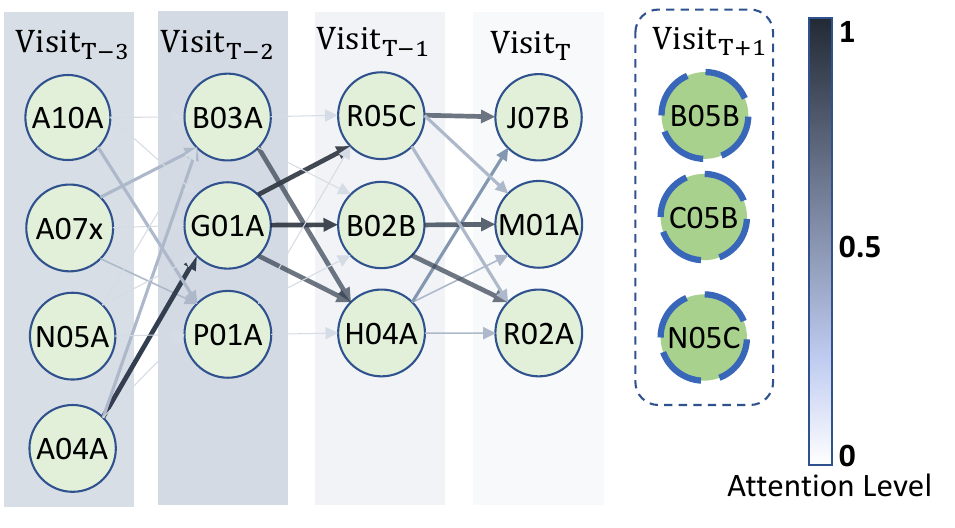}
    \caption{\nth{3} Intent Temporal Path}
  \end{subfigure}
  \begin{subfigure}[h]{0.23\textwidth}
    \includegraphics[width=\textwidth]{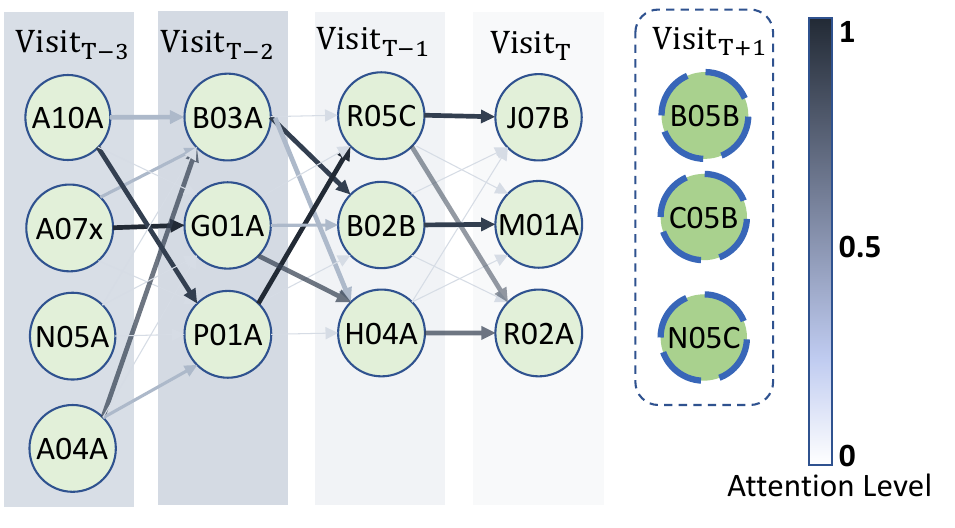}
    \caption{\nth{4} Intent Temporal Path}
  \end{subfigure}
\vspace{0.2cm}
  \caption{Illustration of the attention values and temporal paths generated by the proposed method, using a sample patient from the MIMIC-III dataset. The highlight behind each visit shows the visit-instance attention.}
  \label{interpretability}
\end{figure}

\hspace{2pt} The Cross-Visit Transformer employs an attention matrix to capture temporal paths between two consecutive time steps. In addition to cross-visit attention, the aggregation and prediction layer contains an interpretable visit attention mechanism through recurrent neural networks, showing the influence of each visit on the output prediction. To demonstrate the interpretability of the proposed method, we select one patient from the MIMIC-III dataset and illustrate the attention values for four different intents. As depicted in Figure \ref{interpretability}, each prescription is connected to another one with a temporal path, and the intensity of the colors reflects the connections' strength. 

In the first intent, the sequence of dependencies A04A $\rightarrow$ P01A $\rightarrow$ R05C $\rightarrow$ R02A is one of the influential relations. According to the medical literature, A04A (antiemetics) is associated with P01A (Agents against protozoal diseases) \cite{athavale2020antiemetic}, and R05C (cough suppressants) correlates with R02A (throat preparations)\cite{sultana2016cough}. In addition to temporal paths, the visit-instance attention mechanism highlights the significance of the first two visits compared to the latter ones due to The strong correlation between N05A (antipsychotics) N05C (hypnotics and sedatives) \cite{charney2006hypnotics}, as well as B03A (iron preparations) with B05B (I.V. solutions) \cite{auerbach2008intravenous}.
For the second intent, one of the important dependencies is A04A $\rightarrow$ B03A $\rightarrow$ B02B $\rightarrow$ M01A, and previous studies support associations between B03A (iron preparations) and M01A (antiinflammatory and antirheumatic) \cite{harshman2016role}.
\section{Related Work}
\textbf{Medication Recommendation:} Since past years with the emergence of Deep Learning models \cite{ajami2023unsupervised, nayebi2023contrastive}, numerous longitudinal methodologies have been developed to improve the embeddings derived from Electronic Health Records (EHR) for various applications, including prescription recommendation systems.   Noteworthy early works include Dr. Agent \cite{gao2020dr} wherein an RNN network employs a policy gradient for adaptive learning, and RETAIN \cite{retain} which contains a dual-RNN network to embed multi-level attention mechanism for both medical codes and visits. The main drawback of these early methodologies is that they were not explicitly developed for the prescription recommendation. Enhancing safety and optimizing integration is achievable through a comprehensive understanding of drug-drug interactions (DDI). GAMENet addresses this issue \cite{gamenet1}, which uses a graph-based network to employ DDI as a loss function and evaluation metric. Other approaches like SafeDrug \cite{safedrug} focus on global message passing for recommending prescriptions based on DDI and molecular structures. MICRON \cite{micron} uses a residual neural network to update patient health histories and MoleRec \cite{molerec} focuses on safety and performance improvement in drug recommendations by considering molecular structures.
\\
\noindent \textbf{User Intent and Sequential Recommendation:} 
Sequential recommender systems utilize the user's past behavior sequence in the recommendation process \cite{kang2018self, wang2023sequential}. Earlier approaches often modeled temporal transformations using Markov Chains. For instance, FPMC \cite{10.1145/1772690.1772773} combined Markov Chains and Matrix Factorization to recommend items based on user interests. With the advent of Deep Learning, a new generation of sequential recommenders has emerged. For example, ASReP \cite{liu2021augmenting} contributes to the data sparsity problem by using a pre-trained Transformer-based method to augment the short sequences.
In addition to sequential recommendation systems, intent-aware systems have been developed to capture users' multiple intentions for enhanced recommendations \cite{liu2020basket, wei2021hierarchical}. For instance, Atten-Mixer \cite{attmixer} considers both concept-view and instance-view for items using the Transformer architecture, and ASLI \cite{tanjim2020attentive} extracts users' multiple intents using a temporal convolutional network for predicting the next item in a sequence. Additionally, contrastive learning has been utilized for distinct intent learning in user preferences \cite{wang2023intent}. For instance, ICL \cite{chen2022intent} employs a latent variable that represents the distribution of users and their intent through clustering and contrastive learning. 
\section{Conclusion}
This paper presents \modelfull{}, a prescription recommendation system with two primary contributions. Firstly, we propose a multi-level transformer-based model designed to extract visit-level and cross-visit dependencies to formulate temporal paths between medical codes. Second, we introduced a novel contrastive learning-based approach for transformers' heads linked to different intents as specialized health profiles to handle a variety of dependency types and achieve more comprehensive embedding learning. \modeln{} outperforms the state-of-the-art healthcare models for prescription recommendations on two real-world datasets, while providing interpretable insights into the decision-making process of clinical practitioners.
\bibliographystyle{ACM-Reference-Format}
\balance
\bibliography{References}


\begin{thebibliography}{46}


\ifx \showCODEN    \undefined \def \showCODEN     #1{\unskip}     \fi
\ifx \showDOI      \undefined \def \showDOI       #1{#1}\fi
\ifx \showISBNx    \undefined \def \showISBNx     #1{\unskip}     \fi
\ifx \showISBNxiii \undefined \def \showISBNxiii  #1{\unskip}     \fi
\ifx \showISSN     \undefined \def \showISSN      #1{\unskip}     \fi
\ifx \showLCCN     \undefined \def \showLCCN      #1{\unskip}     \fi
\ifx \shownote     \undefined \def \shownote      #1{#1}          \fi
\ifx \showarticletitle \undefined \def \showarticletitle #1{#1}   \fi
\ifx \showURL      \undefined \def \showURL       {\relax}        \fi
\providecommand\bibfield[2]{#2}
\providecommand\bibinfo[2]{#2}
\providecommand\natexlab[1]{#1}
\providecommand\showeprint[2][]{arXiv:#2}

\bibitem[Ajami et~al\mbox{.}(2023)]%
        {ajami2023unsupervised}
\bibfield{author}{\bibinfo{person}{Hanieh Ajami}, \bibinfo{person}{Mahsa~Kargar
  Nigjeh}, {and} \bibinfo{person}{Scott~E Umbaugh}.}
  \bibinfo{year}{2023}\natexlab{}.
\newblock \showarticletitle{Unsupervised white matter lesion identification in
  multiple sclerosis (MS) using MRI segmentation and pattern classification: a
  novel approach with CVIPtools}. In \bibinfo{booktitle}{\emph{Applications of
  Digital Image Processing XLVI}}, Vol.~\bibinfo{volume}{12674}. SPIE,
  \bibinfo{pages}{282--287}.
\newblock


\bibitem[Ashton et~al\mbox{.}(2023)]%
        {ashton2023using}
\bibfield{author}{\bibinfo{person}{James~J Ashton}, \bibinfo{person}{Aneurin
  Young}, \bibinfo{person}{Mark~J Johnson}, {and} \bibinfo{person}{R~Mark
  Beattie}.} \bibinfo{year}{2023}\natexlab{}.
\newblock \showarticletitle{Using machine learning to impact on long-term
  clinical care: principles, challenges, and practicalities}.
\newblock \bibinfo{journal}{\emph{Pediatric Research}} \bibinfo{volume}{93},
  \bibinfo{number}{2} (\bibinfo{year}{2023}), \bibinfo{pages}{324--333}.
\newblock


\bibitem[Athavale et~al\mbox{.}(2020)]%
        {athavale2020antiemetic}
\bibfield{author}{\bibinfo{person}{Akshay Athavale}, \bibinfo{person}{Tegan
  Athavale}, {and} \bibinfo{person}{Darren~M Roberts}.}
  \bibinfo{year}{2020}\natexlab{}.
\newblock \showarticletitle{Antiemetic drugs: what to prescribe and when}.
\newblock \bibinfo{journal}{\emph{Australian Prescriber}} \bibinfo{volume}{43},
  \bibinfo{number}{2} (\bibinfo{year}{2020}), \bibinfo{pages}{49}.
\newblock


\bibitem[Auerbach et~al\mbox{.}(2008)]%
        {auerbach2008intravenous}
\bibfield{author}{\bibinfo{person}{Michael Auerbach}, \bibinfo{person}{Dan
  Coyne}, {and} \bibinfo{person}{Harold Ballard}.}
  \bibinfo{year}{2008}\natexlab{}.
\newblock \showarticletitle{Intravenous iron: from anathema to standard of
  care}.
\newblock \bibinfo{journal}{\emph{American journal of hematology}}
  \bibinfo{volume}{83}, \bibinfo{number}{7} (\bibinfo{year}{2008}),
  \bibinfo{pages}{580--588}.
\newblock


\bibitem[Bhoi et~al\mbox{.}(2024)]%
        {refine}
\bibfield{author}{\bibinfo{person}{Suman Bhoi}, \bibinfo{person}{Mong~Li Lee},
  \bibinfo{person}{Wynne Hsu}, {and} \bibinfo{person}{Ngiap~Chuan Tan}.}
  \bibinfo{year}{2024}\natexlab{}.
\newblock \showarticletitle{REFINE: A Fine-Grained Medication Recommendation
  System Using Deep Learning and Personalized Drug Interaction Modeling}.
\newblock \bibinfo{journal}{\emph{Advances in Neural Information Processing
  Systems}}  \bibinfo{volume}{36} (\bibinfo{year}{2024}).
\newblock


\bibitem[Charney et~al\mbox{.}(2006)]%
        {charney2006hypnotics}
\bibfield{author}{\bibinfo{person}{Dennis~S Charney}, \bibinfo{person}{S~John
  Mihic}, {and} \bibinfo{person}{R~Adron Harris}.}
  \bibinfo{year}{2006}\natexlab{}.
\newblock \showarticletitle{Hypnotics and sedatives}.
\newblock \bibinfo{journal}{\emph{The Pharmacologic Basis of Therapeutics. 11th
  ed. Brunton LL, Lazo JS, Parker KL (Eds). New York, McGraw-Hill}}
  (\bibinfo{year}{2006}), \bibinfo{pages}{401--427}.
\newblock


\bibitem[Chen et~al\mbox{.}(2020)]%
        {chen2020simple}
\bibfield{author}{\bibinfo{person}{Ting Chen}, \bibinfo{person}{Simon
  Kornblith}, \bibinfo{person}{Mohammad Norouzi}, {and}
  \bibinfo{person}{Geoffrey Hinton}.} \bibinfo{year}{2020}\natexlab{}.
\newblock \showarticletitle{A simple framework for contrastive learning of
  visual representations}. In \bibinfo{booktitle}{\emph{International
  conference on machine learning}}. PMLR, \bibinfo{pages}{1597--1607}.
\newblock


\bibitem[Chen et~al\mbox{.}(2022)]%
        {chen2022intent}
\bibfield{author}{\bibinfo{person}{Yongjun Chen}, \bibinfo{person}{Zhiwei Liu},
  \bibinfo{person}{Jia Li}, \bibinfo{person}{Julian McAuley}, {and}
  \bibinfo{person}{Caiming Xiong}.} \bibinfo{year}{2022}\natexlab{}.
\newblock \showarticletitle{Intent contrastive learning for sequential
  recommendation}. In \bibinfo{booktitle}{\emph{Proceedings of the ACM Web
  Conference 2022}}. \bibinfo{pages}{2172--2182}.
\newblock


\bibitem[Choi et~al\mbox{.}(2016)]%
        {retain}
\bibfield{author}{\bibinfo{person}{Edward Choi}, \bibinfo{person}{Mohammad~Taha
  Bahadori}, \bibinfo{person}{Jimeng Sun}, \bibinfo{person}{Joshua Kulas},
  \bibinfo{person}{Andy Schuetz}, {and} \bibinfo{person}{Walter Stewart}.}
  \bibinfo{year}{2016}\natexlab{}.
\newblock \showarticletitle{Retain: An interpretable predictive model for
  healthcare using reverse time attention mechanism}.
\newblock \bibinfo{journal}{\emph{Advances in neural information processing
  systems}}  \bibinfo{volume}{29} (\bibinfo{year}{2016}).
\newblock


\bibitem[Choi et~al\mbox{.}(2020)]%
        {gct}
\bibfield{author}{\bibinfo{person}{Edward Choi}, \bibinfo{person}{Zhen Xu},
  \bibinfo{person}{Yujia Li}, \bibinfo{person}{Michael Dusenberry},
  \bibinfo{person}{Gerardo Flores}, \bibinfo{person}{Emily Xue}, {and}
  \bibinfo{person}{Andrew Dai}.} \bibinfo{year}{2020}\natexlab{}.
\newblock \showarticletitle{Learning the graphical structure of electronic
  health records with graph convolutional transformer}. In
  \bibinfo{booktitle}{\emph{Proceedings of the AAAI conference on artificial
  intelligence}}, Vol.~\bibinfo{volume}{34}. \bibinfo{pages}{606--613}.
\newblock


\bibitem[Gao et~al\mbox{.}(2020)]%
        {gao2020dr}
\bibfield{author}{\bibinfo{person}{Junyi Gao}, \bibinfo{person}{Cao Xiao},
  \bibinfo{person}{Lucas~M Glass}, {and} \bibinfo{person}{Jimeng Sun}.}
  \bibinfo{year}{2020}\natexlab{}.
\newblock \showarticletitle{Dr. Agent: Clinical predictive model via mimicked
  second opinions}.
\newblock \bibinfo{journal}{\emph{Journal of the American Medical Informatics
  Association}} \bibinfo{volume}{27}, \bibinfo{number}{7}
  (\bibinfo{year}{2020}), \bibinfo{pages}{1084--1091}.
\newblock


\bibitem[Harshman and Shea(2016)]%
        {harshman2016role}
\bibfield{author}{\bibinfo{person}{Stephanie~G Harshman} {and}
  \bibinfo{person}{M~Kyla Shea}.} \bibinfo{year}{2016}\natexlab{}.
\newblock \showarticletitle{The role of vitamin K in chronic aging diseases:
  inflammation, cardiovascular disease, and osteoarthritis}.
\newblock \bibinfo{journal}{\emph{Current nutrition reports}}
  \bibinfo{volume}{5} (\bibinfo{year}{2016}), \bibinfo{pages}{90--98}.
\newblock


\bibitem[Johnson et~al\mbox{.}(2016)]%
        {MIMIC3}
\bibfield{author}{\bibinfo{person}{Alistair~EW Johnson}, \bibinfo{person}{Tom~J
  Pollard}, \bibinfo{person}{Lu Shen}, \bibinfo{person}{Li-wei~H Lehman},
  \bibinfo{person}{Mengling Feng}, \bibinfo{person}{Mohammad Ghassemi},
  \bibinfo{person}{Benjamin Moody}, \bibinfo{person}{Peter Szolovits},
  \bibinfo{person}{Leo Anthony~Celi}, {and} \bibinfo{person}{Roger~G Mark}.}
  \bibinfo{year}{2016}\natexlab{}.
\newblock \showarticletitle{MIMIC-III, a freely accessible critical care
  database}.
\newblock \bibinfo{journal}{\emph{Scientific data}} \bibinfo{volume}{3},
  \bibinfo{number}{1} (\bibinfo{year}{2016}), \bibinfo{pages}{1--9}.
\newblock


\bibitem[Kang and McAuley(2018)]%
        {kang2018self}
\bibfield{author}{\bibinfo{person}{Wang-Cheng Kang} {and}
  \bibinfo{person}{Julian McAuley}.} \bibinfo{year}{2018}\natexlab{}.
\newblock \showarticletitle{Self-attentive sequential recommendation}. In
  \bibinfo{booktitle}{\emph{2018 IEEE international conference on data mining
  (ICDM)}}. IEEE, \bibinfo{pages}{197--206}.
\newblock


\bibitem[Khosla et~al\mbox{.}(2020)]%
        {khosla2020supervised}
\bibfield{author}{\bibinfo{person}{Prannay Khosla}, \bibinfo{person}{Piotr
  Teterwak}, \bibinfo{person}{Chen Wang}, \bibinfo{person}{Aaron Sarna},
  \bibinfo{person}{Yonglong Tian}, \bibinfo{person}{Phillip Isola},
  \bibinfo{person}{Aaron Maschinot}, \bibinfo{person}{Ce Liu}, {and}
  \bibinfo{person}{Dilip Krishnan}.} \bibinfo{year}{2020}\natexlab{}.
\newblock \showarticletitle{Supervised contrastive learning}.
\newblock \bibinfo{journal}{\emph{Advances in neural information processing
  systems}}  \bibinfo{volume}{33} (\bibinfo{year}{2020}),
  \bibinfo{pages}{18661--18673}.
\newblock


\bibitem[Kingma and Ba(2014)]%
        {kingma2014adam}
\bibfield{author}{\bibinfo{person}{Diederik~P Kingma} {and}
  \bibinfo{person}{Jimmy Ba}.} \bibinfo{year}{2014}\natexlab{}.
\newblock \showarticletitle{Adam: A method for stochastic optimization}.
\newblock \bibinfo{journal}{\emph{arXiv preprint arXiv:1412.6980}}
  (\bibinfo{year}{2014}).
\newblock


\bibitem[Koniew(2020)]%
        {koniew2020classification}
\bibfield{author}{\bibinfo{person}{Marek Koniew}.}
  \bibinfo{year}{2020}\natexlab{}.
\newblock \showarticletitle{Classification of the User's Intent Detection in
  Ecommerce systems-Survey and Recommendations.}
\newblock \bibinfo{journal}{\emph{International Journal of Information
  Engineering \& Electronic Business}} \bibinfo{volume}{12},
  \bibinfo{number}{6} (\bibinfo{year}{2020}).
\newblock


\bibitem[Liu et~al\mbox{.}(2022)]%
        {liu2022development}
\bibfield{author}{\bibinfo{person}{Kang Liu}, \bibinfo{person}{Xiangzhou
  Zhang}, \bibinfo{person}{Weiqi Chen}, \bibinfo{person}{SL Alan},
  \bibinfo{person}{John~A Kellum}, \bibinfo{person}{Michael~E Matheny},
  \bibinfo{person}{Steven~Q Simpson}, \bibinfo{person}{Yong Hu}, {and}
  \bibinfo{person}{Mei Liu}.} \bibinfo{year}{2022}\natexlab{}.
\newblock \showarticletitle{Development and validation of a personalized model
  with transfer learning for acute kidney injury risk estimation using
  electronic health records}.
\newblock \bibinfo{journal}{\emph{JAMA Network Open}} \bibinfo{volume}{5},
  \bibinfo{number}{7} (\bibinfo{year}{2022}),
  \bibinfo{pages}{e2219776--e2219776}.
\newblock


\bibitem[Liu et~al\mbox{.}(2021)]%
        {liu2021augmenting}
\bibfield{author}{\bibinfo{person}{Zhiwei Liu}, \bibinfo{person}{Ziwei Fan},
  \bibinfo{person}{Yu Wang}, {and} \bibinfo{person}{Philip~S Yu}.}
  \bibinfo{year}{2021}\natexlab{}.
\newblock \showarticletitle{Augmenting sequential recommendation with
  pseudo-prior items via reversely pre-training transformer}. In
  \bibinfo{booktitle}{\emph{Proceedings of the 44th international ACM SIGIR
  conference on Research and development in information retrieval}}.
  \bibinfo{pages}{1608--1612}.
\newblock


\bibitem[Liu et~al\mbox{.}(2020)]%
        {liu2020basket}
\bibfield{author}{\bibinfo{person}{Zhiwei Liu}, \bibinfo{person}{Xiaohan Li},
  \bibinfo{person}{Ziwei Fan}, \bibinfo{person}{Stephen Guo},
  \bibinfo{person}{Kannan Achan}, {and} \bibinfo{person}{S~Yu Philip}.}
  \bibinfo{year}{2020}\natexlab{}.
\newblock \showarticletitle{Basket recommendation with multi-intent translation
  graph neural network}. In \bibinfo{booktitle}{\emph{2020 IEEE International
  Conference on Big Data (Big Data)}}. IEEE, \bibinfo{pages}{728--737}.
\newblock


\bibitem[Michel et~al\mbox{.}(2019)]%
        {michel2019sixteen}
\bibfield{author}{\bibinfo{person}{Paul Michel}, \bibinfo{person}{Omer Levy},
  {and} \bibinfo{person}{Graham Neubig}.} \bibinfo{year}{2019}\natexlab{}.
\newblock \showarticletitle{Are sixteen heads really better than one?}
\newblock \bibinfo{journal}{\emph{Advances in neural information processing
  systems}}  \bibinfo{volume}{32} (\bibinfo{year}{2019}).
\newblock


\bibitem[Nayebi~Kerdabadi et~al\mbox{.}(2023)]%
        {nayebi2023contrastive}
\bibfield{author}{\bibinfo{person}{Mohsen Nayebi~Kerdabadi},
  \bibinfo{person}{Arya Hadizadeh~Moghaddam}, \bibinfo{person}{Bin Liu},
  \bibinfo{person}{Mei Liu}, {and} \bibinfo{person}{Zijun Yao}.}
  \bibinfo{year}{2023}\natexlab{}.
\newblock \showarticletitle{Contrastive learning of temporal distinctiveness
  for survival analysis in electronic health records}. In
  \bibinfo{booktitle}{\emph{Proceedings of the 32nd ACM International
  Conference on Information and Knowledge Management}}.
  \bibinfo{pages}{1897--1906}.
\newblock


\bibitem[Paszke et~al\mbox{.}(2019)]%
        {pytorch}
\bibfield{author}{\bibinfo{person}{Adam Paszke}, \bibinfo{person}{Sam Gross},
  \bibinfo{person}{Francisco Massa}, \bibinfo{person}{Adam Lerer},
  \bibinfo{person}{James Bradbury}, \bibinfo{person}{Gregory Chanan},
  \bibinfo{person}{Trevor Killeen}, \bibinfo{person}{Zeming Lin},
  \bibinfo{person}{Natalia Gimelshein}, \bibinfo{person}{Luca Antiga},
  {et~al\mbox{.}}} \bibinfo{year}{2019}\natexlab{}.
\newblock \showarticletitle{Pytorch: An imperative style, high-performance deep
  learning library}.
\newblock \bibinfo{journal}{\emph{Advances in neural information processing
  systems}}  \bibinfo{volume}{32} (\bibinfo{year}{2019}).
\newblock


\bibitem[Ranjbar~Kermany et~al\mbox{.}(2022)]%
        {ranjbar2022fair}
\bibfield{author}{\bibinfo{person}{Naime Ranjbar~Kermany},
  \bibinfo{person}{Jian Yang}, \bibinfo{person}{Jia Wu}, {and}
  \bibinfo{person}{Luiz Pizzato}.} \bibinfo{year}{2022}\natexlab{}.
\newblock \showarticletitle{Fair-srs: a fair session-based recommendation
  system}. In \bibinfo{booktitle}{\emph{Proceedings of the Fifteenth ACM
  International Conference on Web Search and Data Mining}}.
  \bibinfo{pages}{1601--1604}.
\newblock


\bibitem[Rendle et~al\mbox{.}(2010)]%
        {10.1145/1772690.1772773}
\bibfield{author}{\bibinfo{person}{Steffen Rendle}, \bibinfo{person}{Christoph
  Freudenthaler}, {and} \bibinfo{person}{Lars Schmidt-Thieme}.}
  \bibinfo{year}{2010}\natexlab{}.
\newblock \showarticletitle{Factorizing personalized Markov chains for
  next-basket recommendation}. In \bibinfo{booktitle}{\emph{Proceedings of the
  19th International Conference on World Wide Web}} (Raleigh, North Carolina,
  USA) \emph{(\bibinfo{series}{WWW '10})}. \bibinfo{publisher}{Association for
  Computing Machinery}, \bibinfo{address}{New York, NY, USA},
  \bibinfo{pages}{811–820}.
\newblock
\showISBNx{9781605587998}
\urldef\tempurl%
\url{https://doi.org/10.1145/1772690.1772773}
\showURL{%
\tempurl}


\bibitem[Shang et~al\mbox{.}(2019)]%
        {gamenet1}
\bibfield{author}{\bibinfo{person}{Junyuan Shang}, \bibinfo{person}{Cao Xiao},
  \bibinfo{person}{Tengfei Ma}, \bibinfo{person}{Hongyan Li}, {and}
  \bibinfo{person}{Jimeng Sun}.} \bibinfo{year}{2019}\natexlab{}.
\newblock \showarticletitle{Gamenet: Graph augmented memory networks for
  recommending medication combination}. In
  \bibinfo{booktitle}{\emph{proceedings of the AAAI Conference on Artificial
  Intelligence}}, Vol.~\bibinfo{volume}{33}. \bibinfo{pages}{1126--1133}.
\newblock


\bibitem[Sultana et~al\mbox{.}(2016)]%
        {sultana2016cough}
\bibfield{author}{\bibinfo{person}{Shahnaz Sultana}, \bibinfo{person}{Andleeb
  Khan}, \bibinfo{person}{Mohammed~M Safhi}, {and} \bibinfo{person}{Hassan~A
  Alhazmi}.} \bibinfo{year}{2016}\natexlab{}.
\newblock \showarticletitle{Cough suppressant herbal drugs: A review}.
\newblock \bibinfo{journal}{\emph{Int. J. Pharm. Sci. Invent}}
  \bibinfo{volume}{5}, \bibinfo{number}{5} (\bibinfo{year}{2016}),
  \bibinfo{pages}{15--28}.
\newblock


\bibitem[Sun et~al\mbox{.}(2022)]%
        {sun2022debiased}
\bibfield{author}{\bibinfo{person}{Hongda Sun}, \bibinfo{person}{Shufang Xie},
  \bibinfo{person}{Shuqi Li}, \bibinfo{person}{Yuhan Chen},
  \bibinfo{person}{Ji-Rong Wen}, {and} \bibinfo{person}{Rui Yan}.}
  \bibinfo{year}{2022}\natexlab{}.
\newblock \showarticletitle{Debiased, Longitudinal and Coordinated Drug
  Recommendation through Multi-Visit Clinic Records}.
\newblock \bibinfo{journal}{\emph{Advances in Neural Information Processing
  Systems}}  \bibinfo{volume}{35} (\bibinfo{year}{2022}),
  \bibinfo{pages}{27837--27849}.
\newblock


\bibitem[Tanjim et~al\mbox{.}(2020)]%
        {tanjim2020attentive}
\bibfield{author}{\bibinfo{person}{Md~Mehrab Tanjim}, \bibinfo{person}{Congzhe
  Su}, \bibinfo{person}{Ethan Benjamin}, \bibinfo{person}{Diane Hu},
  \bibinfo{person}{Liangjie Hong}, {and} \bibinfo{person}{Julian McAuley}.}
  \bibinfo{year}{2020}\natexlab{}.
\newblock \showarticletitle{Attentive sequential models of latent intent for
  next item recommendation}. In \bibinfo{booktitle}{\emph{Proceedings of The
  Web Conference 2020}}. \bibinfo{pages}{2528--2534}.
\newblock


\bibitem[Vaswani et~al\mbox{.}(2017)]%
        {vaswani2017attention}
\bibfield{author}{\bibinfo{person}{Ashish Vaswani}, \bibinfo{person}{Noam
  Shazeer}, \bibinfo{person}{Niki Parmar}, \bibinfo{person}{Jakob Uszkoreit},
  \bibinfo{person}{Llion Jones}, \bibinfo{person}{Aidan~N Gomez},
  \bibinfo{person}{{\L}ukasz Kaiser}, {and} \bibinfo{person}{Illia
  Polosukhin}.} \bibinfo{year}{2017}\natexlab{}.
\newblock \showarticletitle{Attention is all you need}.
\newblock \bibinfo{journal}{\emph{Advances in neural information processing
  systems}}  \bibinfo{volume}{30} (\bibinfo{year}{2017}).
\newblock


\bibitem[Wang et~al\mbox{.}(2023a)]%
        {wang2023sequential}
\bibfield{author}{\bibinfo{person}{Chenyang Wang}, \bibinfo{person}{Weizhi Ma},
  \bibinfo{person}{Chong Chen}, \bibinfo{person}{Min Zhang},
  \bibinfo{person}{Yiqun Liu}, {and} \bibinfo{person}{Shaoping Ma}.}
  \bibinfo{year}{2023}\natexlab{a}.
\newblock \showarticletitle{Sequential recommendation with multiple contrast
  signals}.
\newblock \bibinfo{journal}{\emph{ACM Transactions on Information Systems}}
  \bibinfo{volume}{41}, \bibinfo{number}{1} (\bibinfo{year}{2023}),
  \bibinfo{pages}{1--27}.
\newblock


\bibitem[Wang et~al\mbox{.}(2023b)]%
        {wang2023intent}
\bibfield{author}{\bibinfo{person}{Yuling Wang}, \bibinfo{person}{Xiao Wang},
  \bibinfo{person}{Xiangzhou Huang}, \bibinfo{person}{Yanhua Yu},
  \bibinfo{person}{Haoyang Li}, \bibinfo{person}{Mengdi Zhang},
  \bibinfo{person}{Zirui Guo}, {and} \bibinfo{person}{Wei Wu}.}
  \bibinfo{year}{2023}\natexlab{b}.
\newblock \showarticletitle{Intent-aware recommendation via disentangled graph
  contrastive learning}. In \bibinfo{booktitle}{\emph{Proceedings of the 32th
  international joint conference on artificial intelligence}}.
  \bibinfo{pages}{2343--2351}.
\newblock


\bibitem[Wei et~al\mbox{.}(2021)]%
        {wei2021hierarchical}
\bibfield{author}{\bibinfo{person}{Yinwei Wei}, \bibinfo{person}{Xiang Wang},
  \bibinfo{person}{Xiangnan He}, \bibinfo{person}{Liqiang Nie},
  \bibinfo{person}{Yong Rui}, {and} \bibinfo{person}{Tat-Seng Chua}.}
  \bibinfo{year}{2021}\natexlab{}.
\newblock \showarticletitle{Hierarchical user intent graph network for
  multimedia recommendation}.
\newblock \bibinfo{journal}{\emph{IEEE Transactions on Multimedia}}
  \bibinfo{volume}{24} (\bibinfo{year}{2021}), \bibinfo{pages}{2701--2712}.
\newblock


\bibitem[Wu et~al\mbox{.}(2022)]%
        {wu2022conditional}
\bibfield{author}{\bibinfo{person}{Rui Wu}, \bibinfo{person}{Zhaopeng Qiu},
  \bibinfo{person}{Jiacheng Jiang}, \bibinfo{person}{Guilin Qi}, {and}
  \bibinfo{person}{Xian Wu}.} \bibinfo{year}{2022}\natexlab{}.
\newblock \showarticletitle{Conditional generation net for medication
  recommendation}. In \bibinfo{booktitle}{\emph{Proceedings of the ACM Web
  Conference 2022}}. \bibinfo{pages}{935--945}.
\newblock


\bibitem[Wu et~al\mbox{.}(2019)]%
        {wu2019session}
\bibfield{author}{\bibinfo{person}{Shu Wu}, \bibinfo{person}{Yuyuan Tang},
  \bibinfo{person}{Yanqiao Zhu}, \bibinfo{person}{Liang Wang},
  \bibinfo{person}{Xing Xie}, {and} \bibinfo{person}{Tieniu Tan}.}
  \bibinfo{year}{2019}\natexlab{}.
\newblock \showarticletitle{Session-based recommendation with graph neural
  networks}. In \bibinfo{booktitle}{\emph{Proceedings of the AAAI conference on
  artificial intelligence}}, Vol.~\bibinfo{volume}{33}.
  \bibinfo{pages}{346--353}.
\newblock


\bibitem[Yang et~al\mbox{.}(2023b)]%
        {pyhealth}
\bibfield{author}{\bibinfo{person}{Chaoqi Yang}, \bibinfo{person}{Zhenbang Wu},
  \bibinfo{person}{Patrick Jiang}, \bibinfo{person}{Zhen Lin},
  \bibinfo{person}{Junyi Gao}, \bibinfo{person}{Benjamin Danek}, {and}
  \bibinfo{person}{Jimeng Sun}.} \bibinfo{year}{2023}\natexlab{b}.
\newblock \showarticletitle{{PyHealth}: A Deep Learning Toolkit for Healthcare
  Predictive Modeling}. In \bibinfo{booktitle}{\emph{Proceedings of the 27th
  ACM SIGKDD International Conference on Knowledge Discovery and Data Mining
  (KDD) 2023}}.
\newblock
\urldef\tempurl%
\url{https://github.com/sunlabuiuc/PyHealth}
\showURL{%
\tempurl}


\bibitem[Yang et~al\mbox{.}(2021a)]%
        {yang2021change}
\bibfield{author}{\bibinfo{person}{Chaoqi Yang}, \bibinfo{person}{Cao Xiao},
  \bibinfo{person}{Lucas Glass}, {and} \bibinfo{person}{Jimeng Sun}.}
  \bibinfo{year}{2021}\natexlab{a}.
\newblock \showarticletitle{Change matters: Medication change prediction with
  recurrent residual networks}.
\newblock \bibinfo{journal}{\emph{arXiv preprint arXiv:2105.01876}}
  (\bibinfo{year}{2021}).
\newblock


\bibitem[Yang et~al\mbox{.}(2021b)]%
        {micron}
\bibfield{author}{\bibinfo{person}{Chaoqi Yang}, \bibinfo{person}{Cao Xiao},
  \bibinfo{person}{Lucas Glass}, {and} \bibinfo{person}{Jimeng Sun}.}
  \bibinfo{year}{2021}\natexlab{b}.
\newblock \showarticletitle{Change matters: Medication change prediction with
  recurrent residual networks}.
\newblock \bibinfo{journal}{\emph{arXiv preprint arXiv:2105.01876}}
  (\bibinfo{year}{2021}).
\newblock


\bibitem[Yang et~al\mbox{.}(2021c)]%
        {safedrug}
\bibfield{author}{\bibinfo{person}{Chaoqi Yang}, \bibinfo{person}{Cao Xiao},
  \bibinfo{person}{Fenglong Ma}, \bibinfo{person}{Lucas Glass}, {and}
  \bibinfo{person}{Jimeng Sun}.} \bibinfo{year}{2021}\natexlab{c}.
\newblock \showarticletitle{Safedrug: Dual molecular graph encoders for
  recommending effective and safe drug combinations}.
\newblock \bibinfo{journal}{\emph{arXiv preprint arXiv:2105.02711}}
  (\bibinfo{year}{2021}).
\newblock


\bibitem[Yang et~al\mbox{.}(2023c)]%
        {molerec}
\bibfield{author}{\bibinfo{person}{Nianzu Yang}, \bibinfo{person}{Kaipeng
  Zeng}, \bibinfo{person}{Qitian Wu}, {and} \bibinfo{person}{Junchi Yan}.}
  \bibinfo{year}{2023}\natexlab{c}.
\newblock \showarticletitle{Molerec: Combinatorial drug recommendation with
  substructure-aware molecular representation learning}. In
  \bibinfo{booktitle}{\emph{Proceedings of the ACM Web Conference 2023}}.
  \bibinfo{pages}{4075--4085}.
\newblock


\bibitem[Yang et~al\mbox{.}(2023a)]%
        {yang2023transformehr}
\bibfield{author}{\bibinfo{person}{Zhichao Yang}, \bibinfo{person}{Avijit
  Mitra}, \bibinfo{person}{Weisong Liu}, \bibinfo{person}{Dan Berlowitz}, {and}
  \bibinfo{person}{Hong Yu}.} \bibinfo{year}{2023}\natexlab{a}.
\newblock \showarticletitle{TransformEHR: transformer-based encoder-decoder
  generative model to enhance prediction of disease outcomes using electronic
  health records}.
\newblock \bibinfo{journal}{\emph{Nature Communications}} \bibinfo{volume}{14},
  \bibinfo{number}{1} (\bibinfo{year}{2023}), \bibinfo{pages}{7857}.
\newblock


\bibitem[Yao et~al\mbox{.}(2023)]%
        {yao2023ontology}
\bibfield{author}{\bibinfo{person}{Zijun Yao}, \bibinfo{person}{Bin Liu},
  \bibinfo{person}{Fei Wang}, \bibinfo{person}{Daby Sow}, {and}
  \bibinfo{person}{Ying Li}.} \bibinfo{year}{2023}\natexlab{}.
\newblock \showarticletitle{Ontology-aware prescription recommendation in
  treatment pathways using multi-evidence healthcare data}.
\newblock \bibinfo{journal}{\emph{ACM Transactions on Information Systems}}
  \bibinfo{volume}{41}, \bibinfo{number}{4} (\bibinfo{year}{2023}),
  \bibinfo{pages}{1--29}.
\newblock


\bibitem[Zhang et~al\mbox{.}(2023a)]%
        {attmixer}
\bibfield{author}{\bibinfo{person}{Peiyan Zhang}, \bibinfo{person}{Jiayan Guo},
  \bibinfo{person}{Chaozhuo Li}, \bibinfo{person}{Yueqi Xie},
  \bibinfo{person}{Jae~Boum Kim}, \bibinfo{person}{Yan Zhang},
  \bibinfo{person}{Xing Xie}, \bibinfo{person}{Haohan Wang}, {and}
  \bibinfo{person}{Sunghun Kim}.} \bibinfo{year}{2023}\natexlab{a}.
\newblock \showarticletitle{Efficiently leveraging multi-level user intent for
  session-based recommendation via atten-mixer network}. In
  \bibinfo{booktitle}{\emph{Proceedings of the Sixteenth ACM International
  Conference on Web Search and Data Mining}}. \bibinfo{pages}{168--176}.
\newblock


\bibitem[Zhang et~al\mbox{.}(2023b)]%
        {zhang2023adaptive}
\bibfield{author}{\bibinfo{person}{Yipeng Zhang}, \bibinfo{person}{Xin Wang},
  \bibinfo{person}{Hong Chen}, {and} \bibinfo{person}{Wenwu Zhu}.}
  \bibinfo{year}{2023}\natexlab{b}.
\newblock \showarticletitle{Adaptive disentangled transformer for sequential
  recommendation}. In \bibinfo{booktitle}{\emph{Proceedings of the 29th ACM
  SIGKDD Conference on Knowledge Discovery and Data Mining}}.
  \bibinfo{pages}{3434--3445}.
\newblock


\bibitem[Zhang et~al\mbox{.}(2023c)]%
        {zhang2023knowledge}
\bibfield{author}{\bibinfo{person}{Yingying Zhang}, \bibinfo{person}{Xian Wu},
  \bibinfo{person}{Quan Fang}, \bibinfo{person}{Shengsheng Qian}, {and}
  \bibinfo{person}{Changsheng Xu}.} \bibinfo{year}{2023}\natexlab{c}.
\newblock \showarticletitle{Knowledge-enhanced attributed multi-task learning
  for medicine recommendation}.
\newblock \bibinfo{journal}{\emph{ACM Transactions on Information Systems}}
  \bibinfo{volume}{41}, \bibinfo{number}{1} (\bibinfo{year}{2023}),
  \bibinfo{pages}{1--24}.
\newblock


\bibitem[Zou et~al\mbox{.}(2022)]%
        {zou2022multi}
\bibfield{author}{\bibinfo{person}{Ding Zou}, \bibinfo{person}{Wei Wei},
  \bibinfo{person}{Xian-Ling Mao}, \bibinfo{person}{Ziyang Wang},
  \bibinfo{person}{Minghui Qiu}, \bibinfo{person}{Feida Zhu}, {and}
  \bibinfo{person}{Xin Cao}.} \bibinfo{year}{2022}\natexlab{}.
\newblock \showarticletitle{Multi-level cross-view contrastive learning for
  knowledge-aware recommender system}. In \bibinfo{booktitle}{\emph{Proceedings
  of the 45th International ACM SIGIR Conference on Research and Development in
  Information Retrieval}}. \bibinfo{pages}{1358--1368}.
\newblock


\end{thebibliography}

\end{document}